%% file: main.tex
\documentclass[10pt,twocolumn,letterpaper]{article}
\usepackage[pagenumbers]{cvpr} 
\usepackage{array}      
\usepackage{tabularx}
\usepackage[table]{xcolor}
\usepackage{makecell} 
\usepackage{hhline}
\usepackage{pifont}
\usepackage{booktabs} 
\newcommand{\cmark}{\ding{51}}
\newcommand{\xmark}{\ding{55}}
\usepackage{multirow}
\usepackage{footmisc}

\usepackage[T1]{fontenc}
\usepackage{ragged2e} 
\definecolor{lightgray}{gray}{0.98}
\usepackage{pifont}
\newcolumntype{C}{>{\centering\arraybackslash}p{7mm}}
\newcolumntype{Y}{>{\columncolor{lightgray}\raggedright\arraybackslash}X}


\newcommand{\myratio}{25}

\newcommand{\figref}[1]{Fig.~\ref{#1}}
\newcommand{\tabref}[1]{Tab.~\ref{#1}}

\newcommand{\secref}[1]{Sec.~\ref{#1}}

\newcommand{\Appref}[1]{Appendix~\ref{#1}}

\newcommand{\plainfootnote}[1]{%
  \begingroup
    \renewcommand{\thefootnote}{}
    \footnote{#1}%
  \endgroup
}

\newcommand*{\affaddr}[1]{#1} 
\newcommand*{\affmark}[1][*]{\textsuperscript{#1}}
\newcommand{\authormark}[2][]{%
  \begingroup
  \def\@thefnmark{#1}%
  \footnote{#2}%
  \endgroup
}

\definecolor{cvprblue}{rgb}{0.21,0.49,0.74}
\definecolor{mycolor}{rgb}{0.682,0.914,0.988}
\usepackage[pagebackref,breaklinks,colorlinks,allcolors=cvprblue]{hyperref}

\def\paperID{17223} 
\def\confName{CVPR}
\def\confYear{2026}

\title{IF-Bench: Benchmarking and Enhancing MLLMs for Infrared Images \\with Generative Visual Prompting}

\author{%
Tao Zhang\affmark[1,2*]~~~ 
Yuyang Hong\affmark[1,2*]~~~ 
Yang Xia\affmark[1,2]~~~
Kun Ding\affmark[1]\textsuperscript{$\dagger$}~~~
Zeyu Zhang\affmark[3]~~~  \vspace{3pt} \\
Ying Wang\affmark[1,3]~~~
Shiming Xiang\affmark[1,2]~~~ 
Chunhong Pan\affmark[1,3]~~~ \vspace{3pt} \\
\affaddr{\affmark[1]MAIS, Institute of Automation~~~~~~~~~~} 
\affaddr{\affmark[2]School of Artificial Intelligence, UCAS~~~~~~~~~~} \\
\affaddr{\affmark[3]Research Center of Aerospace Information, Institute of Automation~~~~~~~~~~} \\
\small
}

\begin{document}
\maketitle
\input{sec/0_abstract}    
\input{sec/1_intro}

\input{sec/2_related_work}
\input{sec/3_method}
\input{sec/4_experiments}

\input{sec/5_conclusion}
{
    \small
    \bibliographystyle{ieeenat_fullname}
    \bibliography{main}
}

\input{sec/X_suppl}

\end{document}

%% file: sec/0_abstract.tex
\begin{abstract}
Recent advances in multimodal large language models (MLLMs) have led to impressive progress across various benchmarks. However, their capability in understanding infrared images remains unexplored. To address this gap, we introduce \textbf{IF-Bench}, the first high-quality benchmark designed for evaluating multimodal understanding of infrared images. IF-Bench consists of 499 images sourced from 23 infrared datasets and 680 carefully curated visual question-answer pairs, covering 10 essential dimensions of image understanding. Based on this benchmark, we systematically evaluate over 40 open-source and closed-source MLLMs, employing cyclic evaluation, bilingual assessment, and hybrid judgment strategies to enhance the reliability of the results. Our analysis reveals how model scale, architecture, and inference paradigms affect infrared image comprehension, providing valuable insights for this area. Furthermore, we propose a training-free generative visual prompting (\textbf{GenViP}) method, which leverages advanced image editing models to translate infrared images into semantically and spatially aligned RGB counterparts, thereby mitigating domain distribution shifts. Extensive experiments demonstrate that our method consistently yields significant performance improvements across a wide range of MLLMs. The benchmark and code are available at \url{https://github.com/casiatao/IF-Bench}.
\end{abstract}

%% file: sec/1_intro.tex
\section{Introduction}
\label{sec:intro}

Recently, multimodal large language models (MLLMs) such as GPT-4o \cite{gpt4o}, Gemini-2.5-Flash \cite{gemini2.5}, and Qwen3-VL \cite{qwen3_vl} have achieved remarkable progress in image analysis and understanding, consistently setting new records on various benchmarks \cite{mmbench, mmmu}.\plainfootnote{\textsuperscript{*}Equal contribution. \quad \textsuperscript{$\dagger$}Corresponding author.} However, since these models are primarily trained on natural images, existing evaluations mainly focus on natural scenes, leaving their understanding ability on out-of-domain data, such as infrared images, largely unexplored. Infrared imaging offers superior visibility under low illumination and adverse weather conditions, making it widely used in applications such as surveillance \cite{llvip, flir} and aerial monitoring \cite{dronevehicle}. This naturally raises a critical question: \textit{How well can current MLLMs understand infrared images?} Previous studies \cite{infrared_llava, irgpt} have undertaken preliminary explorations in this direction, but the narrow task coverage, absence of human calibration, and limited model choices leave the actual infrared understanding capability of mainstream MLLMs still unclear.

\begin{figure}[t]
\centering
\includegraphics[width=0.8\linewidth]{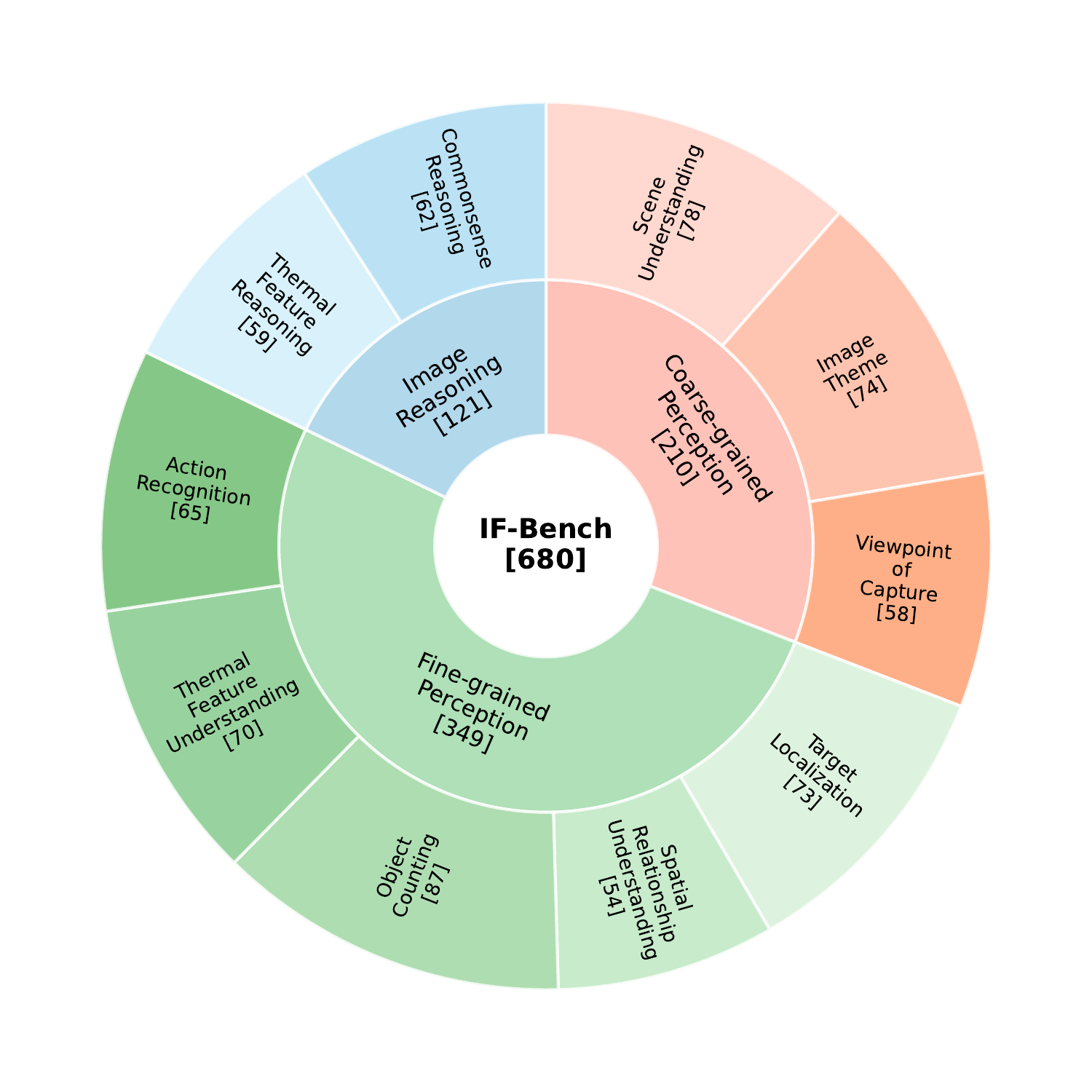}
\caption{Distribution of questions across dimensions in IF-Bench.}
\label{fig:dim_dist}
\vspace{-2mm}
\end{figure}

To address this gap, we propose \textbf{IF-Bench}, a high-quality and systematic benchmark for evaluating multimodal understanding of infrared images. As illustrated in \figref{fig:dim_dist}, we first decompose infrared image understanding into three major tasks: coarse-grained perception, fine-grained perception, and image reasoning, which are further divided into \textbf{10} dimensions to comprehensively cover diverse application scenarios. Subsequently, 1,166 infrared images were selected from 23 datasets, and 4,628 visual question–answer (VQA) pairs were constructed through a hybrid process of manual annotation and automatic generation. After a rigorous two-stage human filtering and calibration process, we obtain the final benchmark consisting of \textbf{499} infrared images and \textbf{680} high-quality VQA pairs.

\begin{figure}[t]
\centering
\includegraphics[width=1.0\linewidth]{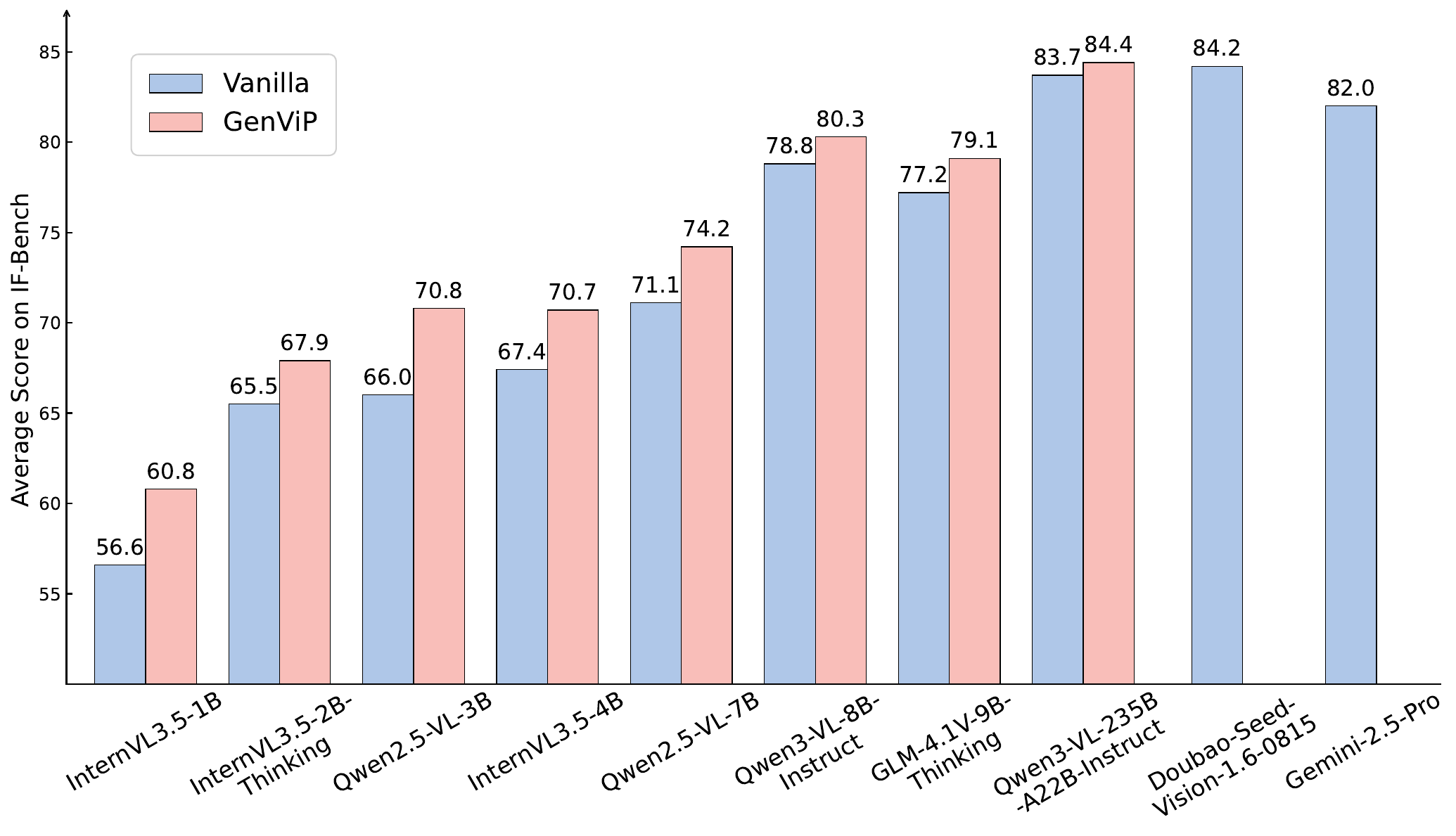}
\vspace{-6mm}
\caption{The performance of GenViP on IF-Bench.}
\label{fig:genvip_hist}
\vspace{-4mm}
\end{figure}

Based on IF-Bench, we comprehensively evaluate over \textbf{40} open-source and closed-source MLLMs with different architectures, scales, and reasoning paradigms, including Qwen2.5-VL \cite{qwen2.5_vl}, Qwen3-VL \cite{qwen3_vl}, InternVL3.5 \cite{internvl3.5}, Gemini-2.5-Pro \cite{gemini2.5}, and Doubao-Seed-Vision \cite{doubao-vision}. To ensure fair and robust evaluation, we design multiple strategies, including the unified evaluation prompt, circular evaluations, bilingual (English and Chinese) assessments, and hybrid correctness judgments. Experimental results exhibit several key findings: (1) increasing model scale consistently improves infrared understanding performance; (2) Mixture-of-Experts (MoE) architectures achieve a better trade-off between accuracy and inference efficiency; (3) The thinking mode enhances performance on \textit{thermal feature understanding and reasoning} dimensions, but reduces accuracy in fine-grained perception tasks; and (4) Open-source models exhibit comparable performance to closed-source models. Overall, existing MLLMs still struggle with comprehending fine details in infrared images. The primary reasons lie partly in the inherent representation limitations of the models themselves and partly in the distribution shift of the input domain: since most MLLMs are primarily trained on RGB images, feeding infrared inputs often introduces substantial domain mismatch, which may lead to performance degradation. Direct adaptation through supervised fine-tuning is challenging due to the scarcity of high-quality infrared–text data and the need for model-specific adjustments.

To overcome these challenges, we propose a training-free \textbf{Gen}erative \textbf{Vi}sual \textbf{P}rompting (\textbf{GenViP}) method. GenViP leverages an image editing model to convert infrared images into semantically and spatially aligned RGB counterparts, which are then jointly fed with the original infrared inputs. This design preserves infrared-specific information while effectively reducing domain gaps. The method requires no paired infrared–text data or model fine-tuning, making it applicable to arbitrary MLLMs. Experimental results indicate that using existing editing models can effectively enable GenViP to enhance infrared image understanding. However, open-source models such as Qwen-Edit-2509 \cite{qwen_edit} still underperform closed-source ones like Seedream 4.0 \cite{seedream4} and Gemini-2.5-Flash \cite{gemini2.5}. To further boost GenViP's performance and practical utility with open-source editing models, we fine-tune Qwen-Edit-2509 \cite{qwen_edit} on 50,000 RGB-T image pairs curated from over 300,000 candidates, allowing it to surpass the aforementioned state-of-the-art (SOTA) closed-source editing models. Extensive results demonstrate that GenViP consistently improves infrared image understanding performance across diverse models, achieving up to \textbf{7\%} relative performance gain on IF-Bench and even outperforming closed-source understanding models like Doubao-Seed-Vision-1.6-250815 \cite{doubao-vision} and Gemini-2.5-Pro \cite{gemini2.5}, as shown in \figref{fig:genvip_hist}.

The main contributions of this work are as follows:
\begin{itemize}
    \item A high-quality benchmark, IF-Bench, for infrared image understanding is constructed, which fills a critical gap in this field and will be publicly released.
    \item A comprehensive and reliable evaluation covering over 40 models, along with in-depth analyses, is presented, providing a solid foundation for future research.
    \item A training-free Generative Visual Prompting (GenViP) method is proposed to effectively enhance the infrared understanding capability of various multimodal models.
\end{itemize}

%% file: sec/2_related_work.tex
\section{Related Work}
\label{sec:formatting}
\input{tables/bench_details}

\noindent\textbf{MLLM.}
The rapid growth of large language models (LLMs) has driven the development of MLLMs. Early works like Flamingo \cite{flamingo} and LLaVA \cite{llava} integrate pre-trained visual encoders with frozen language models for cross-modal alignment. Recent advances further improve visual reasoning and complex semantic understanding. Qwen2.5-VL \cite{qwen2.5_vl} introduces dynamic resolution processing for multi-scale adaptation. LLaVA-OneVision-1.5 \cite{llava_onevision1.5} provides an open framework and a large-scale pre-training dataset, achieving competitive performance across diverse tasks. InternVL3.5 \cite{internvl3.5} leverages cascade reinforcement learning to strengthen visual comprehension and reasoning. Qwen3-VL \cite{qwen3_vl} employs the thinking mode and MoE design to enhance understanding and reasoning in complex tasks. In parallel, closed-source models such as GPT \cite{gpt4o}, Gemini \cite{gemini2.5}, and Doubao \cite{doubao} also advance rapidly, demonstrating strong capabilities in diverse multimodal scenarios. Nevertheless, existing MLLMs still remain limited in infrared image understanding. Previous works like Infrared-LLaVA \cite{infrared_llava} and IRGPT \cite{irgpt} attempt modality-specific adaptation via fine-tuning, but face challenges including scarce image-text data, complex adaptation pipelines, and difficulties in keeping pace with rapidly evolving MLLMs. In contrast, our GenViP requires no training or image-text data and can be directly applied to any MLLM, offering superior convenience and generalization.

\noindent\textbf{Multimodal Understanding Benchmark.}
Systematic benchmarks are essential for quantifying and comparing model performance. MM-Vet \cite{mmvet} and MM-Bench \cite{mmbench} cover a wide range of tasks, including visual understanding and spatial reasoning. MMMU \cite{mmmu} focuses on domain-specific knowledge and high-level reasoning, challenging models to perform tasks akin to those faced by human experts. MVBench \cite{mvbench} transforms static tasks into dynamic ones to evaluate models’ temporal perception and cognitive abilities. MMSI-Bench \cite{mmsi_bench} examines spatial reasoning, scene reconstruction, and spatial transformation capabilities using over 120,000 multi-view images. CAPTURe \cite{capture} assesses the ability to handle occlusion by requiring reasoning about the spatial relationships of partially hidden objects. For infrared image understanding, some works \cite{infrared_llava,irgpt} construct evaluation sets automatically from existing annotated datasets. However, their benchmarks suffer from limited task coverage, a lack of manual verification, and a narrow set of evaluated models, making them insufficient for assessing the infrared understanding abilities of current MLLMs. On the contrary, our proposed IF-Bench is a high-quality benchmark designed with diverse tasks and carefully curated questions. It has been systematically evaluated on more than 40 mainstream MLLMs, providing a robust foundation for future research.

%% file: tables/bench_details.tex
\begin{table*}[t!]
\caption{\textbf{Dimensions and corresponding examples of IF-Bench.} Examples with images are listed in Appendix \textcolor{cvprblue}{A}.}
\vspace{-3mm}
\label{tab:ben_details}
\footnotesize
\renewcommand{\arraystretch}{1.2} 
\setlength{\tabcolsep}{6pt} 

\noindent
\begin{tabularx}{\textwidth}{
  >{\centering\arraybackslash}m{1.8cm}
  | >{\centering\arraybackslash}m{1.9cm}
  | Y
}
\hline
\textbf{Task Category} & \textbf{Dimension} & \makecell[c]{\textbf{Examples}} \\
\hline

\multirow{5}{*}{\makecell[c]{Coarse-grained \\ Perception}}
 &
\makecell[c]{Scene \\ Understanding} &
\makecell{\textit{\textcolor{blue}{What type of environment does this image depict?}}\\(A) Rural road. \ (B) Urban highway. \ (C) Suburban neighborhood. \ (D) Industrial area.} \\

\hhline{~--}

 &
\makecell[c]{Image \\ Theme} &
\makecell{\textit{\textcolor{blue}{What is the most likely theme of this infrared image?}} \\ (A) Traffic surveillance. \ (B) Wildlife monitoring. \ (C) Security surveillance. \ (D) Agricultural monitoring.} \\

\hhline{~--}

 &
\makecell[c]{Viewpoint \\ of Capture} &
\makecell{\textit{\textcolor{blue}{Compared to people in the elevator, from which viewpoint was this image captured?}} \\
(A) Top-down. \ (B) Frontal. \ (C) Side-view. \ (D) Bottom-up.}\\
\hline

\multirow{9}{*}{\makecell[c]{Fine-grained \\ Perception}} &
\makecell[c]{Target \\ Localization} &
\makecell{\textit{\textcolor{blue}{What are the coordinates of the person closest to the straight pole in the image?}}\\ \textit{\textcolor{blue}{(Format: (Target Center X, Target Center Y, Target Width, Target Height))}} \\
(A) (0.4, 0.8, 0.1, 0.3). \ (B) (0.54, 0.9, 0.06, 0.2). \ (C) (0.6, 0.7, 0.05, 0.15). \ (D) (0.5, 0.95, 0.08, 0.25).} \\
\hhline{~--}

 &
 Spatial Relationship Understanding &
\makecell{\textit{\textcolor{blue}{What is the spatial relationship between the two cars in the image?}} \\
(A) The car on the right is behind the car on the left. \ (B) The cars are side by side. \\(C) The car on the right is in front of the car on the left. \ (D) The cars are not visible.} \\
\hhline{~--}

 &
Object Counting &
\makecell{\textit{\textcolor{blue}{How many people are visible in the image?}} \\
(A) Between 20 and 30. \ (B) Between 10 and 20. \ (C) Less than 10. \ (D) More than 30.} \\
\hhline{~--}

 &
Thermal Feature Understanding &
\makecell{\textit{\textcolor{blue}{Which area in the image appears to have the highest thermal activity?}} \\ (A) The crowd of people. \ (B) The trees on the right. \ (C) The ground in the foreground. \ (D) The sky above.} \\
\hhline{~--}

 &
Action Recognition &
\makecell{\textit{\textcolor{blue}{What action is the animal in the image performing?}} \\
(A) Running. \ (B) Walking. \ (C) Jumping. \ (D) Standing still.} \\
\hline

\multirow{3}{*}{\makecell[c]{Image \\ Reasoning}} &

Thermal Feature Reasoning & 
\makecell{\textit{\textcolor{blue}{What could be the reason for the uniform heat signatures among the individuals?}} \\
(A) They are all performing different actions. (B) They are standing in a cold environment. \\(C) The ambient temperature is very high. (D) They are wearing identical clothing materials.} \\
\hhline{~--}
 &
Commonsense Reasoning &
\makecell{\textit{\textcolor{blue}{Based on the image, what is the most likely purpose of the barrier in the foreground?}} \\
(A) To guide traffic flow. \ (B) To block unauthorized access. \\ (C) To provide shade. \ (D) To serve as a decorative element.} \\

\hline

\end{tabularx}
\vspace{-0.3cm}
\end{table*}

%% file: sec/3_method.tex
\section{IF-Bench}

\begin{figure*}[t]
\centering
\includegraphics[width=1.0\linewidth]{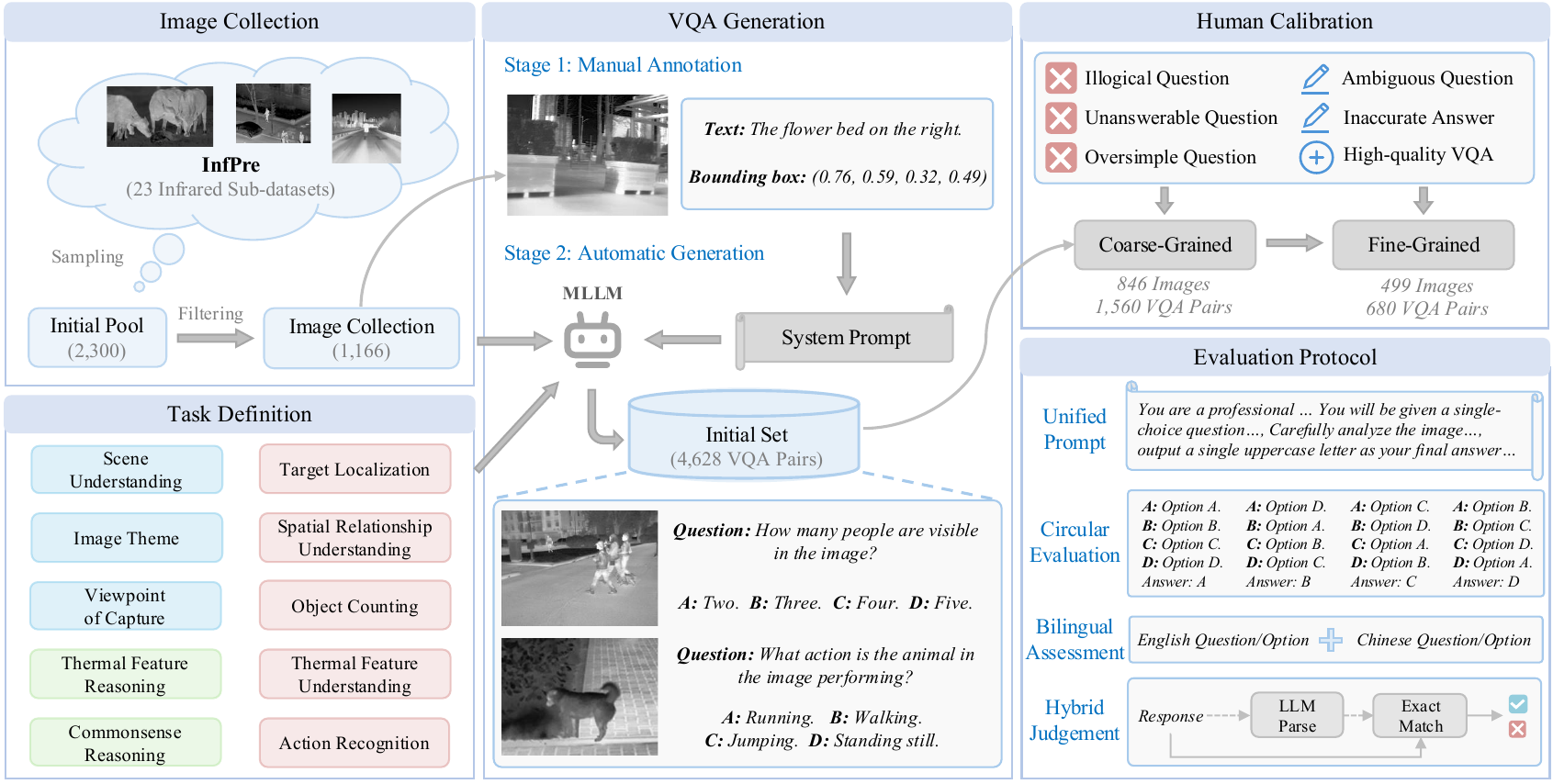}
\vspace{-7mm}
\caption{Construction pipeline and evaluation protocol of IF-Bench.}
\label{fig:pipeline}
\vspace{-2mm}
\end{figure*}

In this section, we first define the tasks involved in infrared image understanding, then present a detailed description of the construction pipeline of IF-Bench, and finally outline its evaluation protocol, as depicted in \figref{fig:pipeline}.

\subsection{Task Definition}
To construct a comprehensive evaluation benchmark for infrared image understanding, we first identify three fundamental tasks: coarse-grained perception, fine-grained perception, and image reasoning. Based on this categorization, we further decompose them into ten dimensions. Detailed definitions of these dimensions are provided below, with corresponding examples shown in \tabref{tab:ben_details}.

\noindent  \textbf{Coarse-grained Perception.} (1) \textit{Scene Understanding}: Identify the general environment or scene depicted in the image, such as indoor, outdoor, forest, or highway. (2) \textit{Image Theme}: Determine the primary content or application context of the image, such as aided driving, wildlife monitoring, or security surveillance. (3) \textit{Viewpoint of Capture}: Identify the viewpoint from which the image was captured, such as top-down, frontal, or side-view.

\noindent \textbf{Fine-grained Perception.} (1) \textit{Target Localization}: Identify the spatial location of a target within the image, including precise localization using a bounding box and rough location, such as upper-left or center. (2) \textit{Spatial Relationship Understanding}: Recognize the spatial relationship between multiple objects in the image, such as front/back or left/right. (3) \textit{Object Counting}: Count the number of objects present in the image. (4) \textit{Thermal Feature Understanding}: Assess thermal variations across image regions, such as detecting heat sources or thermal leaks. (5) \textit{Action Recognition}: Determine whether a target is active and identify its action based on infrared contours.  

\noindent \textbf{Image Reasoning.} (1) \textit{Thermal Feature Reasoning}: Infer potential causes of temperature changes using observed thermal patterns. (2) \textit{Commonsense Reasoning}: Deduce the function or intended use of objects based on image content and infer actions that the objects are likely to perform.

\subsection{Construction Pipeline}
\label{sec:const_pipe}
\noindent \textbf{Image Collection.}
We collect infrared images from InfPre \cite{unip}, a large-scale pre-training dataset that integrates 23 infrared datasets. Specifically, we randomly sample 100 images from each dataset, yielding an initial pool of 2,300 images. To ensure high visual fidelity, images with either width or height below 200 pixels are first excluded, followed by a manual quality assessment to further refine the image pool. Through this two-stage filtering process, we obtain a collection of 1,166 high-quality infrared images.

\noindent \textbf{VQA Generation.}
To facilitate the accurate assessment, we adopt the form of visual question answering with well-defined, deterministic answers. We employ a two-stage procedure to construct the initial set of VQA pairs. (1) \textbf{Manual Annotation:} To provide ground-truth bounding boxes for the \textit{target localization} dimension, we randomly sample 100 images from the image collection. For each image, objects are selected and annotated with a concise textual description and a precise bounding box, while target-free images are excluded. This results in a final set of 61 annotated images. (2) \textbf{Automatic Generation:} Building upon the curated image set, we establish an automated question-answer generation pipeline. Given annotated images with their bounding box and textual description, as well as unannotated images, Qwen2.5-VL-72B \cite{qwen2.5_vl} is prompted to generate at most four single-choice questions per image, each accompanied by four options and the ground-truth answer. The specific system prompt is illustrated in \Appref{sec:appendix_b}. Following this pipeline, we obtain 4,628 VQA pairs.

\noindent \textbf{Human Calibration.}
Considering the potential hallucinations introduced in automatic generation, we apply a coarse-to-fine, two-stage manual calibration process. The calibration follows several key criteria: (1) \textit{Plausibility Assessment}: Remove ill-formed or logically inconsistent questions. (2) \textit{Ambiguity Resolution}: Revise ambiguously phrased questions. For instance, clarifying the observer’s viewpoint in the \textit{spatial relationship understanding} dimension. (3) \textit{Answerability Evaluation}: Exclude questions for whose answers cannot be reliably inferred from the image content.
(4) \textit{Answer Verification}: Correct inaccurate or mismatched answers.
(5) \textit{Difficulty Adjustment}: Filter out highly repetitive or overly simplistic questions to ensure balanced difficulty.
(6) \textit{Data Augmentation}: Introduce additional high-quality VQA pairs grounded in images. The two stages both adhere to the above criteria, while the fine-grained filtering stage is performed by domain experts in infrared imaging, who apply more rigorous quality standards.

Through the above three steps, we obtain the final \textbf{IF-Bench}, which comprises \textbf{499} infrared images and \textbf{680} VQA pairs. Each question is provided in both Chinese and English. The question distribution is shown in \figref{fig:dim_dist}. The final image set still maintains a relatively uniform distribution across the 23 infrared datasets, as analyzed in \Appref{sec:appendix_a}. The order of options is randomly shuffled so that the correct answers are evenly distributed among options A-D. 

\subsection{Evaluation Protocol}
\label{sec:eval_protocol}
To ensure the reliability and robustness of evaluation results, we design several evaluation strategies for IF-Bench, as depicted in \figref{fig:pipeline}. (1) \textbf{Unified Prompt:} All models are evaluated under an identical system prompt and instructed to output only the correct answer for each question. The specific prompt is provided in \Appref{sec:appendix_b}. (2) \textbf{Circular Evaluation:} To mitigate positional bias, we follow the practice in MM-Bench \cite{mmbench}. Specifically, for each question, the four options and the correct answer are cyclically permuted. Each model is evaluated on all permutations, with the final score averaged across them. (3) \textbf{Bilingual Assessment:} Each question is presented in both Chinese and English, with the final score averaged across the two languages. Combined with circular evaluation, each question is evaluated eight times, substantially reducing randomness and improving reliability. (4) \textbf{Hybrid Correctness Judgment:} A hybrid strategy combining exact answer matching and LLM-based parsing is employed to balance accuracy and efficiency in correctness assessment. A model’s response is first checked for an exact match with the ground-truth answer. If no match is found, we utilize Qwen3-7B \cite{qwen3_vl} to extract the answer from the model’s response without adding any extraneous information, which is then compared to the ground truth. The extraction prompt is listed in \Appref{sec:appendix_b}. This strategy effectively mitigates the impact of non-standard output formats on evaluation accuracy.

\section{Infrared Understanding Enhancement}
\subsection{Generative Visual Prompting}
RGB images form the main training data for current MLLMs. However, when these models are applied to infrared images, distribution shifts between training and inference inputs can degrade their visual understanding capabilities. Fine-tuning on infrared understanding data is a straightforward solution but faces several challenges: (1) limited high-quality infrared image–text datasets; (2) high computational and engineering costs for individually fine-tuning each model, considering the rapid progress of MLLMs. (3) possible performance drops on general tasks.

\begin{figure}[t]
\centering
\includegraphics[width=1.0\linewidth]{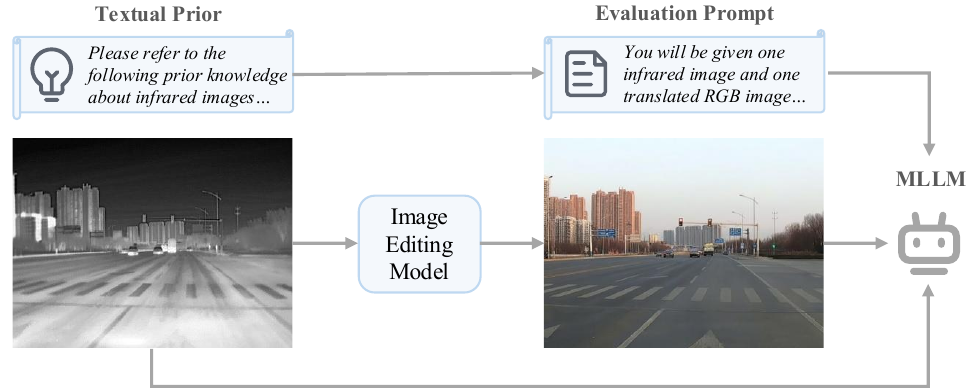}
\vspace{-6mm}
\caption{The illustration of GenViP.}
\label{fig:genvip}
\vspace{-4mm}
\end{figure}

To overcome these limitations, we propose a training-free \textbf{Gen}erative \textbf{Vi}sual \textbf{P}rompting (\textbf{GenViP}) method to improve MLLMs' infrared image understanding.
The key idea is to reduce distribution shifts by directly modifying the model inputs during inference. Specifically, given the advanced capabilities of recent image editing models \cite{seedream4, gemini2.5, qwen_edit}, we translate infrared images into spatially and semantically aligned RGB images. Feeding these translated images into MLLMs effectively aligns the input distribution during inference with that of the training phase. However, since the translated natural images lose thermal information, the models struggle to answer questions about thermal features. Therefore, we adopt a composite-input strategy, feeding both the original infrared image and the translated RGB image into MLLMs, as illustrated in \figref{fig:genvip}. This preserves thermal information while leveraging the model’s strong understanding of RGB images. Consequently, GenViP eliminates the need for high-quality infrared image–text datasets or fine-tuning for each model, making it directly applicable to any existing MLLM. 

To handle the dual-image input, we modify the evaluation prompt accordingly, as shown in \figref{fig:genvip} and \Appref{sec:appendix_b}. Additionally, we incorporate a brief textual description about the characteristics of infrared images into the prompt (\Appref{sec:appendix_b}), namely \textbf{textual prior}, which effectively enhances the model’s ability to interpret infrared thermal features. Extensive experiments subsequently validate the effectiveness of our method in improving the infrared image understanding capabilities of MLLMs.

\begin{table*}[t]
    \centering
    \footnotesize
    \caption{\textbf{Evaluation results of various models on IF-Bench.} Column abbreviations: Avg -- Average score; SU -- Scene Understanding; IT -- Image Theme; VC -- Viewpoint of Capture; TL -- Target Localization; SRU -- Spatial Relationship Understanding; OC -- Object Counting; TFU -- Thermal Feature Understanding; AR -- Action Recognition; TFR -- Thermal Feature Reasoning; CR -- Commonsense Reasoning. Results of closed-source models are marked in \textcolor{gray}{gray}. The highest average scores within each group are marked in \textcolor{cyan}{cyan}.}
    \vspace{-2.5mm}
    \label{tab:if_bench}
    \setlength{\tabcolsep}{1.4mm}{
    \scalebox{0.95}{
    \begin{tabular}{l|>{\centering\arraybackslash}p{10mm}|>{\centering\arraybackslash}p{9mm}>{\centering\arraybackslash}p{9mm}>{\centering\arraybackslash}p{9mm}|CCCCC|>{\centering\arraybackslash}p{10mm}>{\centering\arraybackslash}p{10mm}}
    \toprule
    \multirow{2}{*}{\textbf{Models}} & \multirow{2}{*}{\textbf{Avg}} & \multicolumn{3}{c|}{\textbf{Coarse-grained Perception}} & \multicolumn{5}{c|}{\textbf{Fine-grained Perception}} & \multicolumn{2}{c}{\textbf{Image Reasoning}} \\
    \cmidrule(lr){3-5} \cmidrule(lr){6-10} \cmidrule(lr){11-12}
    &  & SU & IT & VC & TL & SRU & OC & TFU & AR & TFR & CR \\
    \midrule
    InternVL3-1B \cite{internvl3} & 43.0 & 72.4 & 67.4 & 33.2 & 27.9 & 31.0 & 27.7 & 32.9 & 48.3 & 30.3 & 58.9 \\
    InternVL3.5-1B \cite{internvl3.5} & \cellcolor{cyan!\myratio}\textbf{56.6} & 92.9 & 84.1 & 67.2 & 36.3 & 37.5 & 35.2 & 40.0 & 64.2 & 36.9 & 71.6 \\
    InternVL3.5-1B-Thinking \cite{internvl3.5} & 52.5 & 76.6 & 71.1 & 60.6 & 35.3 & 35.2 & 39.8 & 48.2 & 53.7 & 42.2 & 62.7 \\
    \midrule
    InternVL3-2B \cite{internvl3} & 65.6 & 91.7 & 82.1 & 61.9 & 56.3 & 44.7 & 42.1 & 55.7 & 70.0 & 68.2 & 83.1 \\
    InternVL3.5-2B \cite{internvl3.5} & 62.3 & 90.5 & 86.3 & 61.9 & 55.7 & 44.0 & 38.5 & 47.7 & 67.5 & 57.6 & 73.0 \\
    InternVL3.5-2B-Thinking \cite{internvl3.5} & \cellcolor{cyan!\myratio}\textbf{65.5} & 91.0 & 89.5 & 62.5 & 50.2 & 46.5 & 44.1 & 58.9 & 68.5 & 69.9 & 74.0 \\
    \midrule
    Qwen2.5-VL-3B \cite{qwen2.5_vl} & 66.0 & 93.9 & 79.7 & 61.2 & 76.0 & 45.4 & 36.9 & 48.9 & 62.5 & 79.7 & 75.4 \\
    InternVL3.5-4B \cite{internvl3.5} & 67.4 & 92.3 & 85.6 & 69.0 & 58.0 & 55.3 & 41.8 & 58.4 & 65.2 & 71.8 & 76.8 \\
    InternVL3.5-4B-Thinking \cite{internvl3.5} & 71.4 & 92.9 & 88.0 & 76.5 & 61.5 & 58.3 & 43.2 & 64.8 & 65.2 & 86.2 & 77.2 \\
    LLaVA-OneVision-1.5-4B-Instruct \cite{llava_onevision1.5} & 75.2 & 93.3 & 90.2 & 69.4 & 71.9 & 55.1 & 50.6 & 66.6 & 79.0 & 90.7 & 85.3 \\
    Qwen3-VL-4B-Instruct \cite{qwen3_vl} & \cellcolor{cyan!\myratio}\textbf{77.4} & 95.4 & 92.2 & 70.3 & 69.7 & 61.6 & 53.2 & 68.2 & 82.1 & 90.3 & 91.1 \\
    Qwen3-VL-4B-Thinking \cite{qwen3_vl} & 76.8 & 93.9 & 92.6 & 83.2 & 59.8 & 59.7 & 51.3 & 66.4 & 84.0 & 87.7 & 89.7 \\
    \midrule
    Qwen2.5-VL-7B \cite{qwen2.5_vl} & 71.1 & 93.3 & 86.5 & 74.4 & 60.3 & 49.5 & 42.7 & 58.9 & 74.0 & 88.6 & 83.1 \\
    InternVL3-8B \cite{internvl3} & 71.2 & 95.0 & 87.8 & 71.6 & 66.4 & 51.9 & 45.0 & 57.7 & 79.0 & 70.8 & 87.1 \\
    InternVL3.5-8B \cite{internvl3.5} & 69.1 & 92.6 & 79.7 & 66.8 & 66.7 & 52.3 & 45.0 & 61.8 & 67.7 & 82.8 & 75.6 \\
    InternVL3.5-8B-Thinking \cite{internvl3.5} & 72.3 & 92.8 & 85.5 & 65.3 & 67.3 & 58.6 & 47.3 & 65.0 & 71.0 & 91.5 & 78.8 \\
    Keye-VL-1.5-8B \cite{keye_vl1.5} & 71.5 & 85.1 & 82.6 & 70.7 & 77.9 & 61.1 & 42.1 & 60.9 & 71.9 & 80.9 & 82.1 \\
    Keye-VL-1.5-8B-Auto-Thinking \cite{keye_vl1.5} & 71.7 & 85.4 & 84.5 & 72.6 & 73.1 & 59.5 & 43.0 & 61.8 & 73.1 & 81.6 & 82.7 \\
    Keye-VL-1.5-8B-Thinking \cite{keye_vl1.5} & 73.3 & 86.1 & 85.3 & 77.4 & 72.1 & 63.0 & 45.8 & 64.8 & 69.2 & 89.0 & 80.6 \\
    LLaVA-OneVision-1.5-8B-Instruct \cite{llava_onevision1.5} & 75.9 & 95.0 & 89.7 & 70.9 & 73.1 & 54.2 & 54.6 & 62.0 & 79.2 & 91.9 & 88.3 \\
    Qwen3-VL-8B-Instruct \cite{qwen3_vl} & \cellcolor{cyan!\myratio}\textbf{78.8} & 93.9 & 93.1 & 79.7 & 74.1 & 65.5 & 60.2 & 68.6 & 75.8 & 88.3 & 88.9 \\
    Qwen3-VL-8B-Thinking \cite{qwen3_vl} & 76.5 & 94.6 & 91.2 & 74.8 & 62.0 & 65.3 & 58.0 & 69.3 & 76.7 & 89.8 & 83.1 \\
    GLM-4.1V-9B-Thinking \cite{glm4} & 77.2 & 94.4 & 92.1 & 77.8 & 67.8 & 59.7 & 55.0 & 69.3 & 77.7 & 89.4 & 88.9 \\
    \midrule
    InternVL3-14B \cite{internvl3} & 73.8 & 91.3 & 91.0 & 73.7 & 64.9 & 59.3 & 48.4 & 65.9 & 74.2 & 82.4 & 86.3 \\
    InternVL3.5-14B \cite{internvl3.5} & 73.6 & 94.1 & 92.1 & 66.4 & 74.0 & 56.9 & 48.6 & 70.2 & 67.3 & 89.9 & 76.8 \\
    InternVL3.5-14B-Thinking \cite{internvl3.5} & \cellcolor{cyan!\myratio}\textbf{74.6} & 93.9 & 93.1 & 71.3 & 69.9 & 58.6 & 46.0 & 72.7 & 69.6 & 92.6 & 78.6 \\
    \midrule
    InternVL3.5-20B-A4B \cite{internvl3.5} & 69.7 & 90.5 & 89.4 & 68.8 & 63.9 & 55.1 & 42.2 & 60.9 & 67.9 & 76.1 & 81.9 \\
    InternVL3.5-20B-A4B-Thinking \cite{internvl3.5} & 69.8 & 90.7 & 91.2 & 63.4 & 58.0 & 56.0 & 40.7 & 66.8 & 64.2 & 87.1 & 80.0 \\
    InternVL3.5-30B-A3B \cite{internvl3.5} & 74.4 & 94.9 & 88.2 & 73.7 & 75.0 & 59.0 & 45.0 & 69.0 & 68.1 & 85.8 & 85.7 \\
    InternVL3.5-30B-A3B-Thinking \cite{internvl3.5} & 75.4 & 93.4 & 90.0 & 74.1 & 69.0 & 62.0 & 48.4 & 71.4 & 68.8 & 92.4 & 84.5 \\
    Qwen3-VL-30B-A3B-Instruct \cite{qwen3_vl} & \cellcolor{cyan!\myratio}\textbf{82.3} & 96.2 & 95.3 & 83.4 & 79.6 & 62.5 & 61.9 & 69.1 & 85.2 & 98.1 & 91.9 \\
    Qwen3-VL-30B-A3B-Thinking \cite{qwen3_vl} & 79.6 & 95.2 & 94.4 & 84.9 & 68.8 & 60.6 & 60.2 & 67.9 & 77.5 & 95.6 & 90.7 \\
    \midrule
    Qwen2.5-VL-32B \cite{qwen2.5_vl} & 75.4 & 94.9 & 89.7 & 75.6 & 73.5 & 59.7 & 45.1 & 63.0 & 71.5 & 96.0 & 84.5 \\
    InternVL3-38B \cite{internvl3} & 78.9 & 93.8 & 92.4 & 76.3 & 76.0 & 61.1 & 58.3 & 69.1 & 78.5 & 94.3 & 89.7 \\
    InternVL3.5-38B \cite{internvl3.5} & 79.0 & 93.4 & 93.2 & 77.8 & 80.5 & 57.6 & 59.6 & 70.0 & 77.7 & 91.5 & 88.7 \\
    InternVL3.5-38B-Thinking \cite{internvl3.5} & \cellcolor{cyan!\myratio}\textbf{80.1} & 94.1 & 94.4 & 78.4 & 75.2 & 62.5 & 65.8 & 76.3 & 72.5 & 92.4 & 89.7 \\
    \midrule
    Qwen2.5-VL-72B \cite{qwen2.5_vl} & 78.1 & 93.9 & 93.1 & 80.6 & 72.4 & 64.1 & 50.0 & 66.8 & 76.9 & 94.9 & 88.3 \\
    InternVL3-78B \cite{internvl3} & \text{80.2} & 94.2 & 94.1 & 78.7 & 74.1 & 62.3 & 61.9 & 73.6 & 80.4 & 95.1 & 87.9 \\
    GLM-4.5V-106B-A12B-Thinking \cite{glm4} & \cellcolor{cyan!\myratio}\textbf{80.3} & 95.2 & 95.3 & 80.6 & 82.0 & 61.8 & 57.6 & 69.1 & 82.1 & 93.9 & 85.7 \\
    \midrule
    Qwen3-VL-235B-A22B-Instruct \cite{qwen3_vl} & 83.7 & 95.8 & 94.9 & 84.5 & 83.2 & 65.3 & 68.4 & 71.6 & 84.0 & 97.5 & 92.1 \\
    Qwen3-VL-235B-A22B-Thinking \cite{qwen3_vl} & 82.8 & 95.5 & 95.4 & 81.9 & 79.6 & 67.1 & 64.7 & 75.4 & 80.2 & 94.9 & 92.9 \\
    InternVL3.5-241B-A28B \cite{internvl3.5} & 83.9 & 95.5 & 95.3 & 82.1 & 88.0 & 66.7 & 70.7 & 73.8 & 82.3 & 95.8 & 89.1 \\
    InternVL3.5-241B-A28B-Thinking \cite{internvl3.5} & 83.9 & 95.4 & 95.9 & 81.5 & 85.3 & 65.3 & 69.4 & 81.1 & 82.3 & 93.9 & 88.9 \\
    \rowcolor{gray!15}Doubao-Seed-1.6-251015 \cite{doubao} & 79.9 & 92.5 & 92.9 & 83.4 & 77.6 & 67.6 & 54.3 & 74.6 & 78.3 & 91.1 & 86.9 \\
    \rowcolor{gray!15}Doubao-Seed-Vision-1.6-250815 \cite{doubao-vision} & \cellcolor{cyan!\myratio}\textbf{84.2} & 92.3 & 91.9 & 87.9 & 87.7 & 68.5 & 63.2 & 78.6 & 83.1 & 96.6 & 91.9 \\
    \rowcolor{gray!15}Gemini-2.5-Flash \cite{gemini2.5} & 79.8 & 94.2 & 94.6 & 75.9 & 76.0 & 67.6 & 62.6 & 70.7 & 71.5 & 93.2 & 91.1 \\
    \rowcolor{gray!15}Gemini-2.5-Pro \cite{gemini2.5} & 82.0 & 94.9 & 89.9 & 85.3 & 80.8 & 63.9 & 63.2 & 76.4 & 80.0 & 94.1 & 91.9 \\
    \bottomrule
\end{tabular}}}
\vspace{-2mm}
\end{table*}

\subsection{Editing Model Optimization}
Our preliminary investigations show that the translation quality of editing models strongly affects GenViP's performance: closed-source editing models such as Seedream 4.0 \cite{seedream4} and Gemini-2.5-Flash \cite{gemini2.5} outperform open-source models like Qwen-Edit-2509 \cite{qwen_edit}. Considering the cost of using closed-source editing models, we further optimize GenViP for open-source models to improve its practicality.

\begin{table}[t]
    \centering
    \footnotesize
    \caption{\textbf{Correlation coefficients between model scale and scores on IF-Bench.} We first compute the correlation coefficients within each model family, and then average the results across families. For MoE models, we use their total parameter count.}
    \vspace{-2mm}
    \label{tab:pearson_coefficient}
    \setlength{\tabcolsep}{1.0mm}{
    \scalebox{0.97}{
    \begin{tabular}{l|c|cccccccccc}
    \toprule
    & Avg & SU & IT & VC & TL & SRU & OC & TFU & AR & TFR & CR \\
    \midrule
    Corr & 0.76 & 0.42 & 0.67 & 0.69 & 0.58 & 0.70 & 0.86 & 0.77 & 0.63 & 0.63 & 0.62 \\ 
    \bottomrule
\end{tabular}}}
\vspace{-2mm}
\end{table}

Initially, we collect over 370,000 RGB-T image pairs and apply rigorous data filtering and sampling as follows:
(1) \textit{Data Source Screening}: Manually remove sources where infrared and RGB images are misaligned.
(2) \textit{Resolution filtering}: Discard pairs with a width or height below 200 pixels.
(3) \textit{Pairing Quality Filtering}: First, remove pairs with excessively low brightness. Next, compute the Canny edges \cite{canny} for both the infrared and RGB images, filtering out pairs whose Dice coefficient falls below a threshold. Finally, using Qwen2.5-VL-32B \cite{qwen2.5_vl} to assess the pairing quality of the remaining data.
(4) \textit{Feature Extraction and Deduplication}: Extract visual features using DINOv3 \cite{dinov3}, and remove pairs whose infrared features closely match those in IF-Bench to prevent data leakage.
(5) \textit{Hierarchical Clustering and Balanced Sampling}: After the above steps, 156,330 high-quality pairs remain. To ensure a balanced distribution across scenes, we perform hierarchical clustering \cite{hir_sample} based on the RGB DINOv3 features, followed by balanced sampling \cite{hir_sample}. Finally, a high-quality dataset of 50,000 RGB-T image pairs is constructed. 

Based on this dataset, we fine-tune Qwen-Edit-2509 \cite{qwen_edit} and obtain \textbf{Qwen-Edit-2509-FT}, which achieves superior infrared-to-RGB translation quality and even outperforms closed-source editing models on IF-Bench.

%% file: sec/4_experiments.tex
\section{Experiments}
\subsection{Evaluation Models.}
We systematically evaluate over 40 mainstream MLLMs on IF-Bench, covering diverse architectures, parameter scales, and reasoning paradigms, including InternVL3 \cite{internvl3}, InternVL3.5 \cite{internvl3.5}, Qwen2.5-VL \cite{qwen2.5_vl}, Qwen3-VL \cite{qwen3_vl}, LLaVA-OneVision \cite{llava_onevision1.5}, Keye-VL-1.5 \cite{keye_vl1.5}, GLM-4.1V \cite{glm4}, GLM-4.5V \cite{glm4}, Gemini-2.5-Pro \cite{gemini2.5}, and Doubao-Seed-Vision-1.6 \cite{doubao-vision}. These models range from 1B to 241B parameters, encompassing both dense and (MoE) architectures, with and without explicit reasoning capabilities. Collectively, they provide a representative and comprehensive snapshot of the current multimodal model landscape.

\subsection{Evaluation Results of IF-Bench.}
The complete evaluation results are presented in \tabref{tab:if_bench}. We conduct an in-depth analysis in the following aspects.

\noindent \textbf{Model Scale and Architecture.}
(1) \textcolor{blue!60}{Significant performance gaps are observed between models of different architectures or series}, particularly for smaller models (<30B). For instance, LLaVA-OneVision-1.5-4B-Instruct and Qwen2.5-VL-3B differ by 9.2 in average score, while Qwen3-VL-8B-Instruct and InternVL3.5-8B are nearly 10.0 points apart. As model scale increases, the performance differences between architectures gradually diminish. Qwen3-VL-235B-A22B-Instruct and InternVL3.5-241B-A28B-Thinking differ by only 0.2 points (83.7 vs 84.2). (2) \textcolor{blue!60}{Smaller models of the Qwen3-VL series demonstrate outstanding performance,} highlighting their practical value for infrared image understanding. For example, Qwen3-VL-8B achieves an overall score of 78.8, surpassing that of Qwen2.5-VL-72B and approaching InternVL3-78B, using only approximately one-tenth of the parameters. (3) \textcolor{blue!60}{MoE models provide a favorable trade-off between performance and inference efficiency.} InternVL3.5-30B-A3B outperforms the fully dense InternVL3.5-14B (74.4 vs 73.6), while only activating fewer than one-quarter of parameters.
(4) To further evaluate the relationship between model size and performance, we compute their Pearson correlation coefficients, as shown in \tabref{tab:pearson_coefficient}. The results indicate that \textcolor{blue!60}{model size is positively correlated with performance across all dimensions}, with particularly strong correlations observed on more challenging tasks, such as \textit{object counting} and \textit{thermal feature understanding}.

\begin{figure}[t]
\centering
\includegraphics[width=1.0\linewidth]{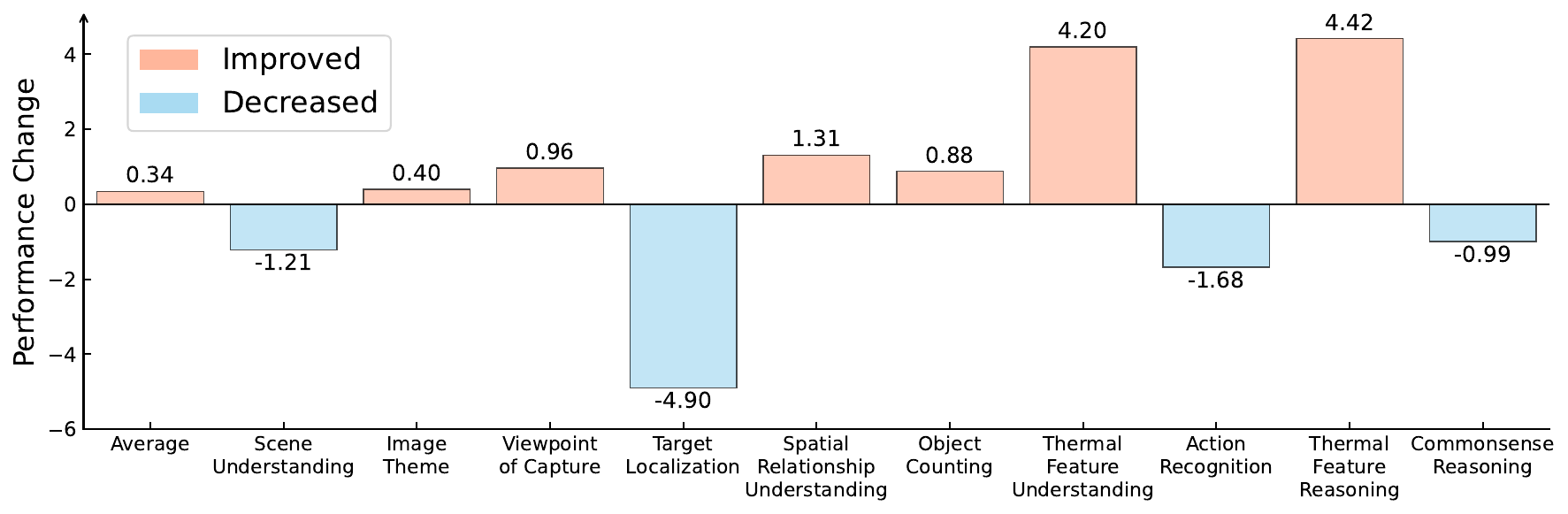}
\vspace{-7mm}
\caption{The average performance change after using thinking.}
\label{fig:thinking}
\vspace{-2mm}
\end{figure}

\begin{figure}[t]
\centering
\includegraphics[width=1.0\linewidth]{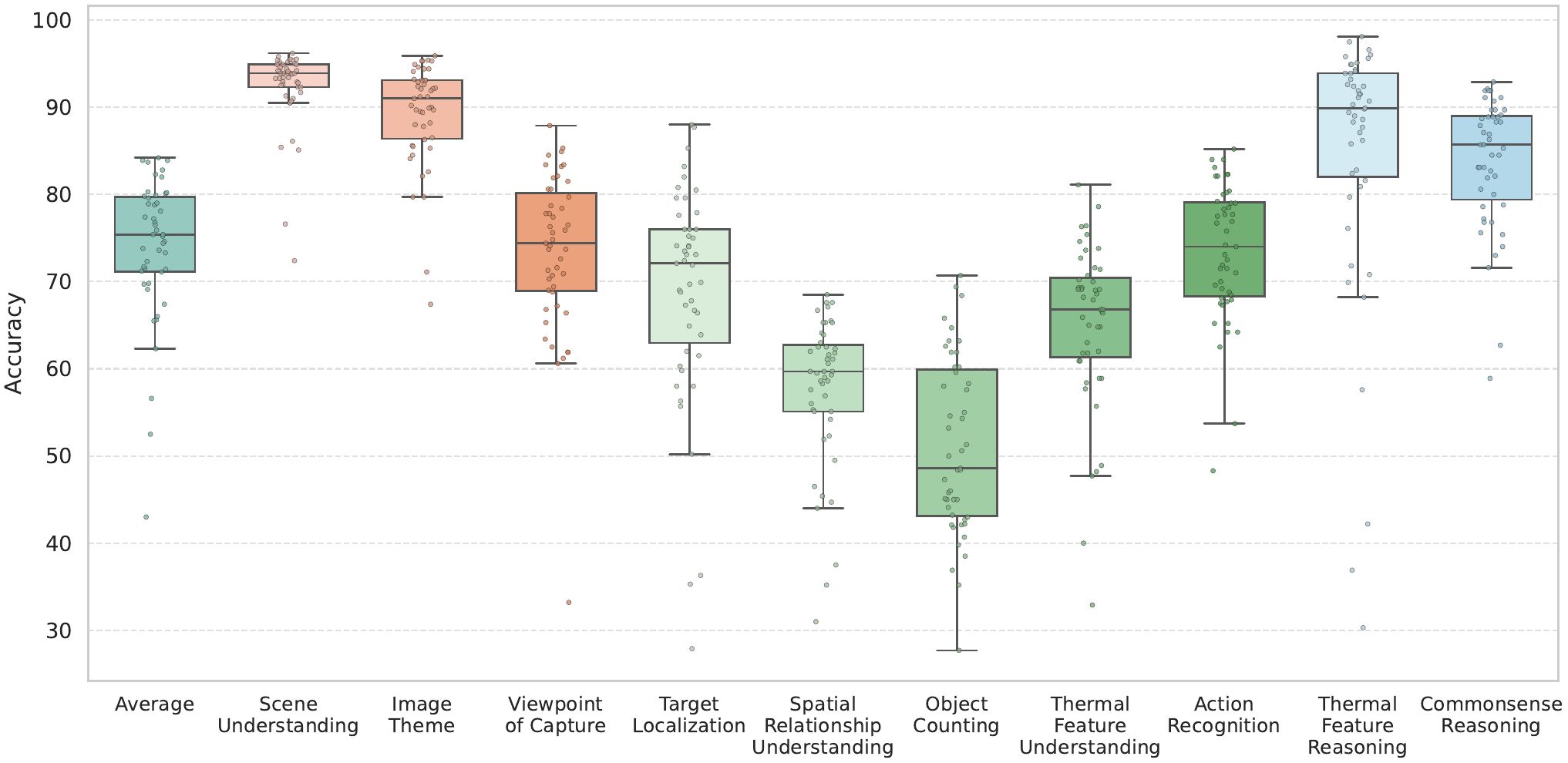}
\vspace{-7mm}
\caption{Box plot of scores across dimensions in IF-Bench.}
\label{fig:if_bench_box_plot}
\vspace{-5mm}
\end{figure}

\noindent\textbf{Thinking vs Non-Thinking.} To explore the impact of the thinking mode, we analyze 14 paired thinking and non-thinking variants in \tabref{tab:if_bench}. Interestingly, the effectiveness of thinking varies considerably across model families. In the Qwen3-VL series, enabling thinking results in a noticeable performance drop, whereas all InternVL3.5 models from 2B to 38B benefit from thinking. As shown in \figref{fig:thinking}, thinking enhances performance in \textit{thermal feature understanding} (+4.2) and \textit{thermal feature reasoning} (+4.42), but reduces accuracy in dimensions such as \textit{target localization} (-4.90) and \textit{action recognition} (-1.68). Overall, the thinking mode yields only a modest average improvement of 0.34. Considering the substantial increase in reasoning overhead it introduces, \textcolor{blue!60}{incorporating thinking is not an effective approach to enhancing infrared image understanding.}

\input{tables/merge_table}

\noindent\textbf{Open-Source vs Closed-Source.} The performance gap between open-source and closed-source models is minimal. Although Doubao-Seed-Vision-1.5-250815 \cite{doubao-vision} achieves the highest score of 84.2, surpassing all open-source models, it exceeds the best open-source model, InternVL3.5-241B-A28B, by only 0.3 points. Moreover, the other three closed-source models, including Gemini-2.5-Pro \cite{gemini2.5}, perform worse than Qwen3-VL-235B-A22B and InternVL3.5-241B-A28B, underscoring that \textcolor{blue!60}{open-source models have significant practical value for infrared image understanding.}

\noindent\textbf{Evaluation Dimension.} 
The accuracy distribution across different dimensions is illustrated in \figref{fig:if_bench_box_plot}. Overall, \textit{scene understanding} and \textit{image theme} in coarse-grained perception tasks, as well as the two reasoning dimensions, are relatively easy, with average scores exceeding 80. In contrast, the five fine-grained perception dimensions and the \textit{viewpoint of capture} dimension exhibit notably lower performance. In particular, for \textit{object counting} and \textit{spatial relationship understanding}, the average accuracies drop below 60, with the highest scores only around 70. These findings indicate that \textcolor{blue!60}{current MLLMs still face challenges in capturing fine-grained details in infrared image understanding.}

\subsection{Results of GenViP}
\noindent\textbf{Main Results.}
As shown in \tabref{tab:if_bench_genvip}, GenViP consistently improves performance across various models. It achieves over 4 points improvement on InternVL3.5-1B and Qwen2.5-VL-3B. Qwen3-VL-8B-Instruct with GenViP matches the performance of InternVL3-78B and GLM-4.5V-106B-A12B-Thinking in \tabref{tab:if_bench}, while using only one-tenth of their parameters. Moreover, Qwen3-VL-235B-A22B-Instruct with GenViP attains a score of 84.4, outperforming all other models in \tabref{tab:if_bench}, including closed-source ones. The consistent performance gains across diverse models highlight the effectiveness and generalization ability of GenViP. However, the improvement diminishes as the model scale increases. This trend may stem from two factors: (1) Larger models already achieve high baseline performance, leaving limited room for further improvement. (2) Larger models may possess stronger robustness to input distribution shifts, reducing their reliance on domain adaptation techniques like GenViP.

\noindent\textbf{Editing Models.}
\tabref{tab:if_bench_edit_model} compares the performance of various image editing models employed in GenViP. All models enable GenViP to effectively enhance infrared image understanding. However, the open-source Qwen-Edit-2509 lags behind top-performing closed-source models such as Seedream 4.0 \cite{seedream4} and Gemini-2.5-Flash \cite{gemini2.5}. After fine-tuning on our high-quality RGB-T paired dataset, Qwen-Edit-2509-FT effectively bridges this performance gap and even surpasses the closed-source models on this specific task. We provide a visual comparison of the translation quality across different models in \Appref{sec:appendix_d}.

\noindent\textbf{Inference Input.} 
\tabref{tab:if_bench_input} examines the effect of different input configurations on infrared image understanding performance. Incorporating the textual prior (b) improves \textit{thermal feature understanding and reasoning} compared with using only infrared images (a), but degrades performance on other dimensions, resulting in negligible overall improvement. Using only the translated RGB image (c) causes substantial drops in \textit{image theme}, \textit{thermal feature understanding and reasoning} due to the loss of thermal information. Nevertheless, by mitigating domain shift, it improves tasks that rely less on infrared-specific cues, such as \textit{spatial relationship understanding}, \textit{object counting}, and \textit{action recognition}. Combining both infrared and RGB images (d) consistently enhances performance across most dimensions. Further adding the textual prior yields additional gains in \textit{thermal understanding and reasoning}, leading to the best overall performance on IF-Bench.

%% file: tables/merge_table.tex
\begin{table*}[t]
\centering
\scriptsize
\begin{minipage}[t]{0.53\textwidth}
    \setlength{\tabcolsep}{0.6mm}
\caption{Performance of GenViP on IF-Bench across various models.}
 \vspace{-3mm}
\label{tab:if_bench_genvip}
\renewcommand{\arraystretch}{1.05} 
\scalebox{0.93}{
\begin{tabular}{lc|l|cccccccccc}
\toprule
Model & GenViP & \makecell[c]{Avg} & SU & IT & VC & TL & SRU & OC & TFU & AR & TFR & CR \\
\midrule

\multirow{2}{*}{InternVL3.5-1B \cite{internvl3.5}} & \xmark & 56.6 & 92.9 & 84.1 & 67.2 & 36.3 & 37.5 & 35.2 & 40.0 & 64.2 & 36.9 & 71.6 \\ 
 & \cellcolor{cyan!15}\cmark & \cellcolor{cyan!15}\textbf{60.8 \textcolor{ForestGreen}{ (+4.2)}} & \cellcolor{cyan!15}94.9 & \cellcolor{cyan!15}85.8 & \cellcolor{cyan!15}67.5 & \cellcolor{cyan!15}34.6 & \cellcolor{cyan!15}41.7 & \cellcolor{cyan!15}38.8 & \cellcolor{cyan!15}34.3 & \cellcolor{cyan!15}69.0 & \cellcolor{cyan!15}60.6 & \cellcolor{cyan!15}80.6 \\
\midrule

\multirow{2}{*}{\makecell[c]{InternVL3.5-2B-\\Thinking \cite{internvl3.5}}} & \xmark & 65.5 & 91.0 & 89.5 & 62.5 & 50.2 & 46.5 & 44.1 & 58.9 & 68.5 & 69.9 & 74.0 \\
& \cellcolor{cyan!15}\cmark & \cellcolor{cyan!15}\textbf{67.9 \textcolor{ForestGreen}{ (+2.4)}} & \cellcolor{cyan!15}95.7 & \cellcolor{cyan!15}87.7 & \cellcolor{cyan!15}68.5 & \cellcolor{cyan!15}49.3 & \cellcolor{cyan!15}51.4 & \cellcolor{cyan!15}35.2 & \cellcolor{cyan!15}53.8 & \cellcolor{cyan!15}76.2 & \cellcolor{cyan!15}83.9 & \cellcolor{cyan!15}77.8 \\
\midrule

\multirow{2}{*}{Qwen2.5-VL-3B \cite{qwen2.5_vl}} & \xmark & 66.0 & 93.9 & 79.7 & 61.2 & 76.0 & 45.4 & 36.9 & 48.9 & 62.5 & 79.7 & 75.4 \\
 & \cellcolor{cyan!15}\cmark & \cellcolor{cyan!15}\textbf{70.8 \textcolor{ForestGreen}{ (+4.8)}} & \cellcolor{cyan!15}97.1 & \cellcolor{cyan!15}82.1 & \cellcolor{cyan!15}68.5 & \cellcolor{cyan!15}76.0 & \cellcolor{cyan!15}44.9 & \cellcolor{cyan!15}40.9 & \cellcolor{cyan!15}51.3 & \cellcolor{cyan!15}78.5 & \cellcolor{cyan!15}87.9 & \cellcolor{cyan!15}81.0 \\
\midrule

\multirow{2}{*}{InternVL3.5-4B \cite{internvl3.5}} & \xmark & 67.4 & 92.3 & 85.6 & 69.0 & 58.0 & 55.3 & 41.8 & 58.4 & 65.2 & 71.8 & 76.8 \\
& \cellcolor{cyan!15}\cmark & \cellcolor{cyan!15}\textbf{70.7 \textcolor{ForestGreen}{ (+2.7)}} & \cellcolor{cyan!15}96.5 & \cellcolor{cyan!15}87.8 & \cellcolor{cyan!15}72.8 & \cellcolor{cyan!15}51.4 & \cellcolor{cyan!15}54.9 & \cellcolor{cyan!15}44.0 & \cellcolor{cyan!15}49.8 & \cellcolor{cyan!15}76.7 & \cellcolor{cyan!15}86.9 & \cellcolor{cyan!15}86.3 \\
\midrule

\multirow{2}{*}{Qwen2.5-VL-7B \cite{qwen2.5_vl}} & \xmark & 71.1 & 93.3 & 86.5 & 74.4 & 60.3 & 49.5 & 42.7 & 58.9 & 74.0 & 88.6 & 83.1 \\
 & \cellcolor{cyan!15}\cmark & \cellcolor{cyan!15}\textbf{74.2 \textcolor{ForestGreen}{ (+3.1)}} & \cellcolor{cyan!15}94.9 & \cellcolor{cyan!15}90.4 & \cellcolor{cyan!15}75.0 & \cellcolor{cyan!15}57.4 & \cellcolor{cyan!15}55.8 & \cellcolor{cyan!15}48.3 & \cellcolor{cyan!15}60.4 & \cellcolor{cyan!15}80.4 & \cellcolor{cyan!15}93.0 & \cellcolor{cyan!15}86.1 \\ 
\midrule

\multirow{2}{*}{\makecell[c]{Qwen3-VL-8B-\\Instruct \cite{qwen3_vl}}} & \xmark & 78.8 & 93.9 & 93.1 & 79.7 & 74.1 & 65.5 & 60.2 & 68.6 & 75.8 & 88.3 & 88.9 \\
 & \cellcolor{cyan!15}\cmark & \cellcolor{cyan!15}\textbf{80.3 \textcolor{ForestGreen}{(+1.5)}} & \cellcolor{cyan!15}95.2 & \cellcolor{cyan!15}94.9 & \cellcolor{cyan!15}78.7 & \cellcolor{cyan!15}68.2 & \cellcolor{cyan!15}66.9 & \cellcolor{cyan!15}59.2 & \cellcolor{cyan!15}70.2 & \cellcolor{cyan!15}82.7 & \cellcolor{cyan!15}96.4 & \cellcolor{cyan!15}90.3 \\
\midrule

\multirow{2}{*}{\makecell[c]{GLM-4.1V-9B-\\Thinking \cite{glm4}}} & \xmark & 77.2 & 94.4 & 92.1 & 77.8 & 67.8 & 59.7 & 55.0 & 69.3 & 77.7 & 89.4 & 88.9 \\
& \cellcolor{cyan!15}\cmark & \cellcolor{cyan!15}\textbf{79.1 \textcolor{ForestGreen}{(+1.9)}} & \cellcolor{cyan!15}96.6 & \cellcolor{cyan!15}93.9 & \cellcolor{cyan!15}72.4 & \cellcolor{cyan!15}69.3 & \cellcolor{cyan!15}62.7 & \cellcolor{cyan!15}61.1 & \cellcolor{cyan!15}70.7 & \cellcolor{cyan!15}80.0 & \cellcolor{cyan!15}95.3 & \cellcolor{cyan!15}88.9 \\
\midrule

\multirow{2}{*}{\makecell[c]{Qwen3-VL-235B-\\A22B-Instruct \cite{qwen3_vl}}} & \xmark & 83.7 & 95.8 & 94.9 & 84.5 & 83.2 & 65.3 & 68.4 & 71.6 & 84.0 & 97.5 & 92.1 \\
 & \cellcolor{cyan!15}\cmark & \cellcolor{cyan!15}\textbf{84.4 \textcolor{ForestGreen}{(+0.7)}} & \cellcolor{cyan!15}96.0 & \cellcolor{cyan!15}96.1 & \cellcolor{cyan!15}85.1 & \cellcolor{cyan!15}83.6 & \cellcolor{cyan!15}66.2 & \cellcolor{cyan!15}66.8 & \cellcolor{cyan!15}74.1 & \cellcolor{cyan!15}85.6 & \cellcolor{cyan!15}97.5 & \cellcolor{cyan!15}92.9 \\
\bottomrule
\end{tabular}}
\end{minipage}
\hfill
\begin{minipage}[t]{0.45\textwidth}
    \setlength{\tabcolsep}{0.6mm} 
\caption{Impact of editing models using Qwen2.5-VL-7B \cite{qwen2.5_vl}.}
\vspace{-3mm}
\label{tab:if_bench_edit_model}

\scalebox{0.92}{
\begin{tabular}{lccccccccccc}
\toprule
Edit Model & Avg & SU & IT & VC & TL & SRU & OC & TFU & AR & TFR & CR \\
\midrule
Vanilla & 71.1 & 93.3 & 86.5 & 74.4 & 60.3 & 49.5 & 42.7 & 58.9 & 74.0 & 88.6 & 83.1 \\
\midrule
\textit{Closed-Source Models} \\
Seedream 4.0 \cite{seedream4} & 73.8 & 94.2 & 89.7 & 77.4 & 56.9 & 53.0 & 50.0 & 60.0 & 79.8 & 92.8 & 84.1 \\
Gemini-2.5-Flash \cite{gemini2.5} & 73.6 & 95.7 & 91.6 & 79.5 & 57.4 & 48.6 & 50.7 & 58.8 & 79.4 & 92.8 & 82.1 \\
\midrule
\textit{Open-Source Models} \\
Qwen-Edit-2509 \cite{qwen_edit} & 72.7 & 92.6 & 88.5 & 75.4 & 61.3 & 51.4 & 46.6 & 62.5 & 72.7 & 92.8 & 83.5 \\
\rowcolor{cyan!15}Qwen-Edit-2509-FT & \textbf{74.2} & 94.9 & 90.4 & 75.0 & 57.4 & 55.8 & 48.3 & 60.4 & 80.4 & 93.0 & 86.1 \\ 
\bottomrule
\end{tabular}}

\vspace{2.5mm}
\setlength{\tabcolsep}{0.65mm} 
\caption{Impact of inference input with Qwen2.5-VL-7B \cite{qwen2.5_vl}, where RGB images are generated by Qwen-Edit-2509-FT. Abbreviations: IF -- Infrared Image. Text --Textual Prior.}
\vspace{-3mm}
\label{tab:if_bench_input}
\scalebox{0.92}{
\begin{tabular}{llccccccccccc}
\toprule
 & Inference Input & Avg & SU & IT & VC & TL & SRU & OC & TFU & AR & TFR & CR \\
\midrule
(a) & IF & 71.1 & 93.3 & 86.5 & 74.4 & \textbf{60.3} & 49.5 & 42.7 & 58.9 & 74.0 & 88.6 & 83.1 \\
(b) & IF + Text & 71.2 & 92.5 & 85.8 & 74.8 & 55.1 & 51.6 & 42.7 & \textbf{62.5} & 71.9 & \textbf{93.0} & 81.9 \\
(c) & RGB & 69.2 & \textbf{96.3} & 76.4 & 69.2 & 58.0 & 54.2 & 46.1 & 50.0 & \textbf{80.8} & 77.8 & 83.7 \\
(d) & IF + RGB & 73.2 & 96.2 & \textbf{91.2} & 73.5 & 58.0 & 55.6 & \textbf{49.3} & 55.0 & 80.4 & 87.3 & 85.7 \\
\rowcolor{cyan!15}(e) & IF + RGB + Text & \textbf{74.2} & 94.9 & 90.4 & \textbf{75.0} & 57.4 & \textbf{55.8} & 48.3 & 60.4 & 80.4 & \textbf{93.0} & \textbf{86.1} \\
\bottomrule
    \end{tabular}}
    
\end{minipage}
\end{table*}

%% file: sec/5_conclusion.tex
\section{Conclusion}
In this work, we present IF-Bench, a comprehensive benchmark for evaluating multimodal understanding of infrared images, along with several evaluation strategies designed to ensure reliable assessment. Through an extensive evaluation of over 40 MLLMs, we uncover several key findings: (1) model size and architecture substantially influence performance; (2) the thinking mode does not effectively improve overall results; (3) open-source and closed-source models exhibit comparable performance; and (4) current MLLMs still perform poorly on fine-grained infrared image understanding. We believe these insights can inspire future advances in this domain. Furthermore, we introduce GenViP to enhance the infrared understanding capability of existing MLLMs. By leveraging generative models to produce information-rich visual prompts, GenViP significantly improves model generalization, offering evidence that generation can effectively facilitate visual understanding. We hope our work provides new perspectives and inspiration for the broader multimodal research community.

\noindent\textbf{Limitations and Future Work.} The current IF-Bench contains a relatively limited number of images and questions, and does not cover some more challenging task types. In future work, we will further expand and diversify the benchmark to keep pace with the rapid progress of MLLMs and explore more methods to improve infrared understanding.

%% file: sec/X_suppl.tex
\clearpage
\appendix
\setcounter{page}{1}
\maketitlesupplementary
\renewcommand\thesection{\Alph{section}}
\setcounter{figure}{6}
\setcounter{table}{6}

We illustrate additional example cases and the image distribution of IF-Bench in \Appref{sec:appendix_a}. All prompts used in the paper are listed in \Appref{sec:appendix_b}. \Appref{sec:appendix_c} details the training and inference settings for editing models in GenViP. In \Appref{sec:appendix_d}, we conduct further analyses, including comparisons of the language preferences of MLLMs on IF-Bench, the stricter correctness judgement strategy, the connection between our work and Thinking-with-Image, and the translated quality across editing models.

\section{Details of IF-Bench}
\label{sec:appendix_a}
\begin{figure*}[htp]
\centering
\includegraphics[width=1.0\linewidth]{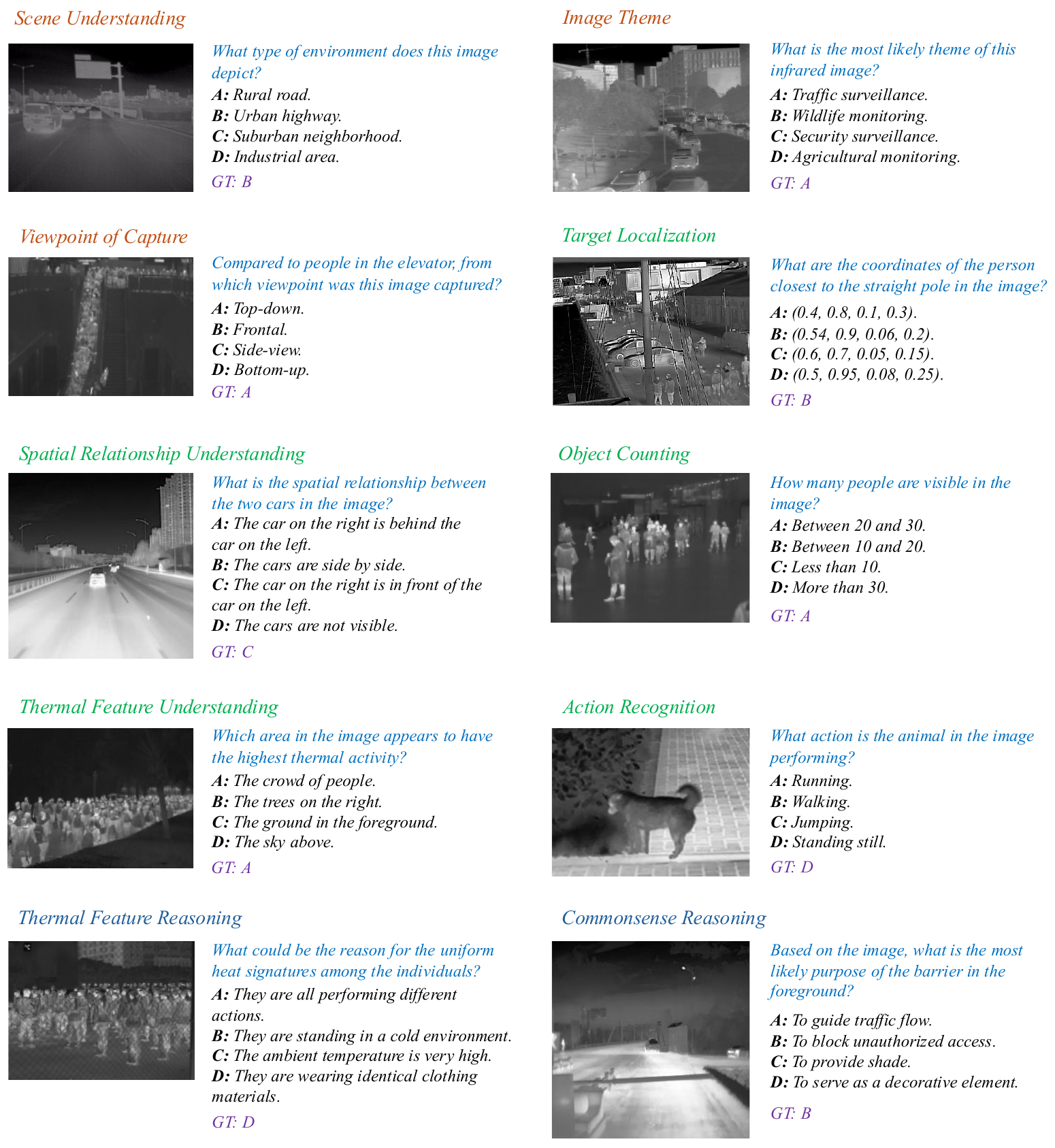}
\vspace{-5mm}
\caption{Example cases for each dimension in IF-Bench, aligned with the questions and options in Tab.~\textcolor{cvprblue}{2}.}
\label{fig:example_case}
\vspace{-3mm}
\end{figure*}

\begin{figure*}[hbt]
\centering
\includegraphics[width=1.0\linewidth]{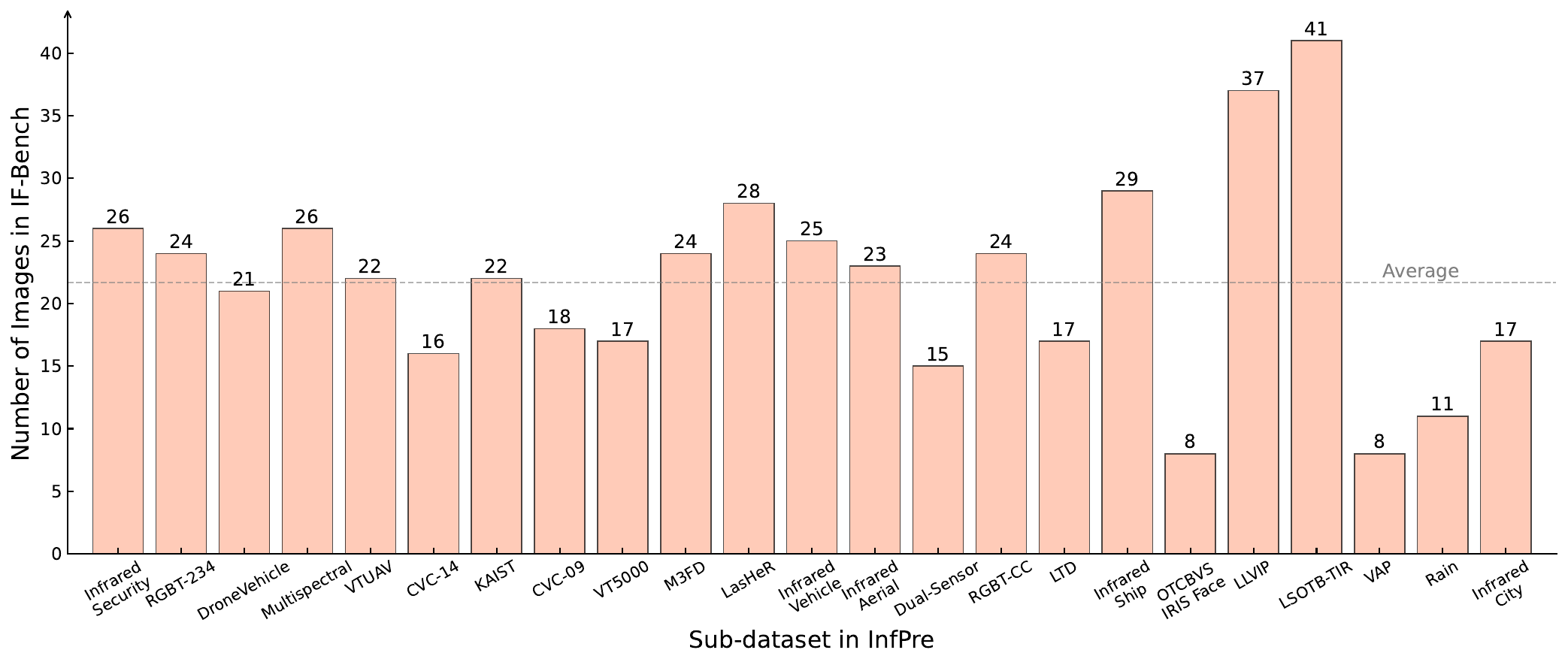}
\vspace{-5mm}
\caption{The distribution images in IF-Bench across various sub-datasets in InfPre \cite{unip}.}
\label{fig:image_dist}
\vspace{-3mm}
\end{figure*}

In this section, we present additional details of IF-Bench. \figref{fig:example_case} illustrates some cases in IF-Bench, with images, questions, options, and ground-truth answers. \figref{fig:image_dist} depicts the image distribution in IF-Bench across 23 sub-datasets in InfPre \cite{unip}. Notably, after manual filtering, the images in IF-Bench remain relatively evenly distributed across the constituent sub-datasets, ensuring comprehensive coverage of diverse scenarios. The datasets OTCBVS-IRIS-FACE \cite{face}, VAP \cite{vap}, and Rain \cite{rainsnow} contain fewer samples primarily because their scenes are comparatively homogeneous.

\begin{figure*}[hbt]
\centering
\includegraphics[width=1.0\linewidth]{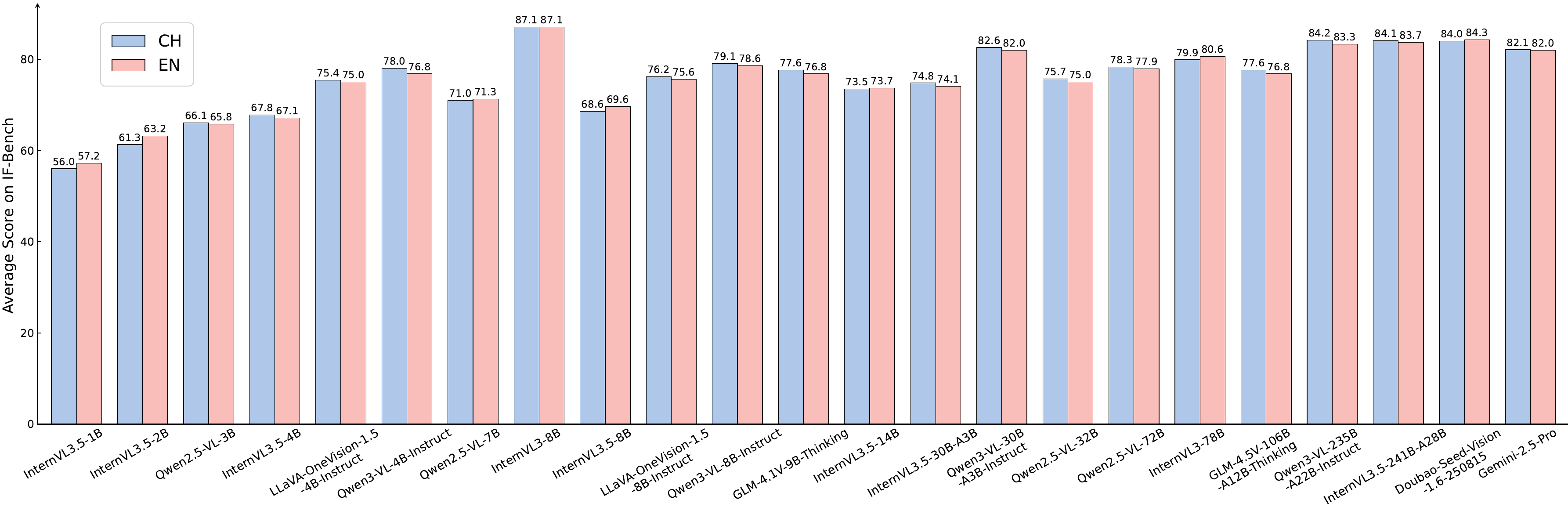}
\vspace{-5mm}
\caption{The language-specific score on IF-Bench of various MLLMs.}
\label{fig:language_pref}
\vspace{-3mm}
\end{figure*}

\section{Details of Prompts}
\label{sec:appendix_b}
\begin{figure}[hbt]
\centering
\includegraphics[width=1.0\linewidth]{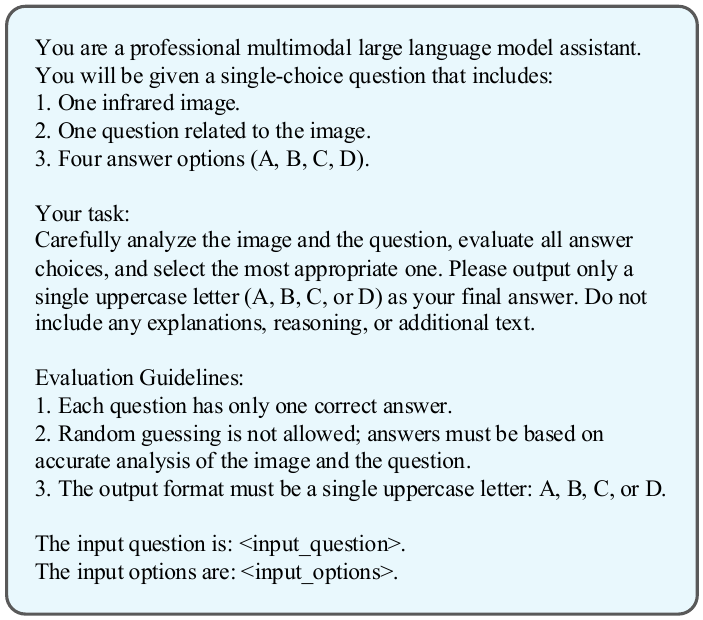}
\vspace{-6mm}
\caption{The unified system prompt when evaluating MLLMs on IF-Bench (English Version).}
\label{fig:eval_prompt}
\vspace{-2mm}
\end{figure}

\begin{figure}[hbt]
\centering
\includegraphics[width=1.0\linewidth]{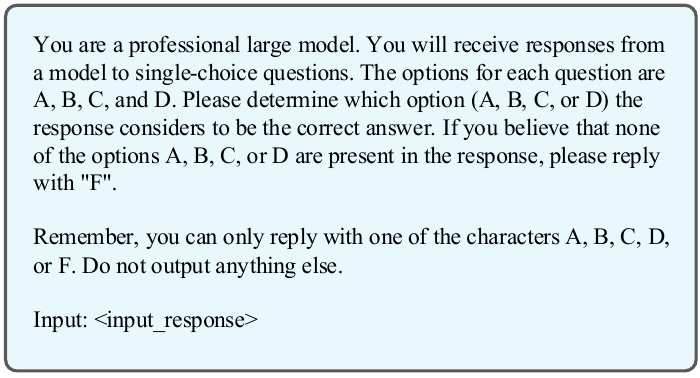}
\vspace{-6mm}
\caption{The system prompt for LLM parsing in \secref{sec:eval_protocol}.}
\label{fig:parse_prompt}
\vspace{-2mm}
\end{figure}

\begin{figure}[hbt]
\centering
\includegraphics[width=1.0\linewidth]{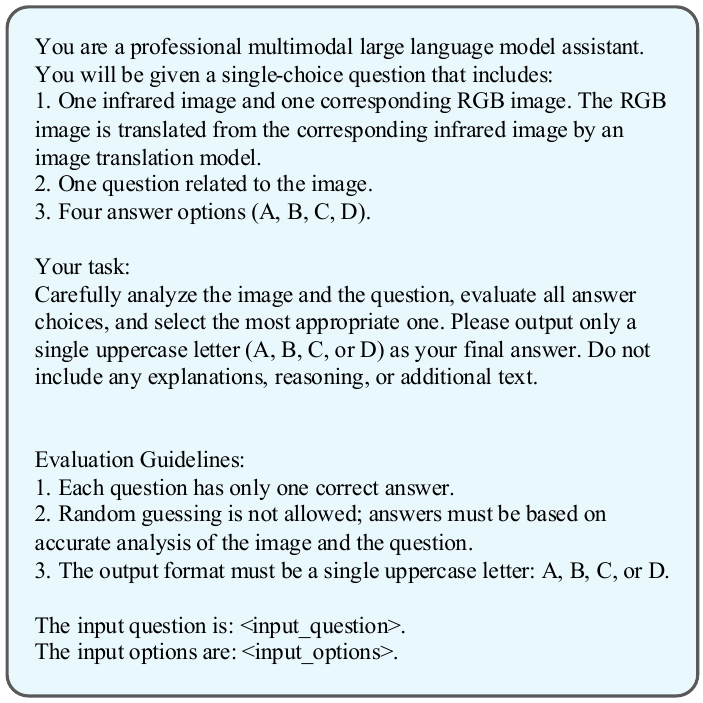}
\vspace{-6mm}
\caption{The evaluation prompt when using the dual-image input in GenViP (English Version).}
\label{fig:eval_prompt_dual}
\vspace{-2mm}
\end{figure}

\begin{figure}[hbt]
\centering
\includegraphics[width=1.0\linewidth]{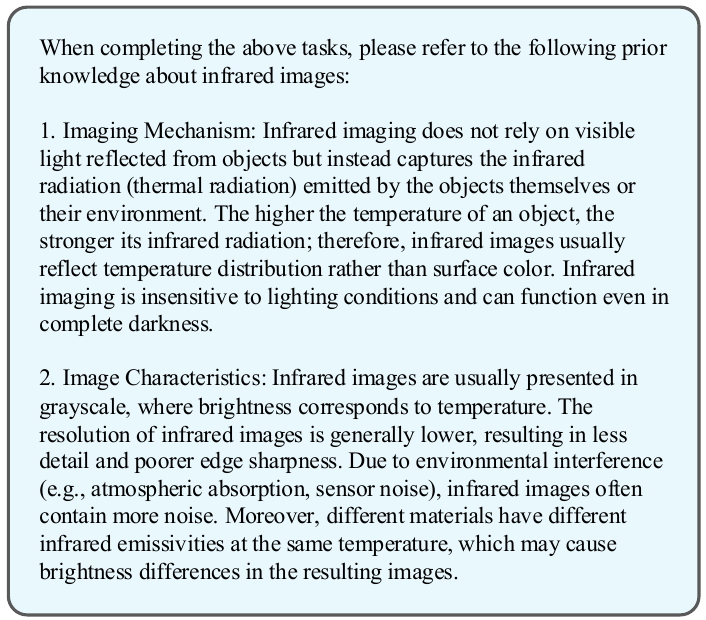}
\vspace{-6mm}
\caption{The textual prior for infrared images (English Version).}
\label{fig:textual_prior}
\vspace{-2mm}
\end{figure}

\begin{figure*}[p]
\centering
\includegraphics[width=0.98\linewidth]{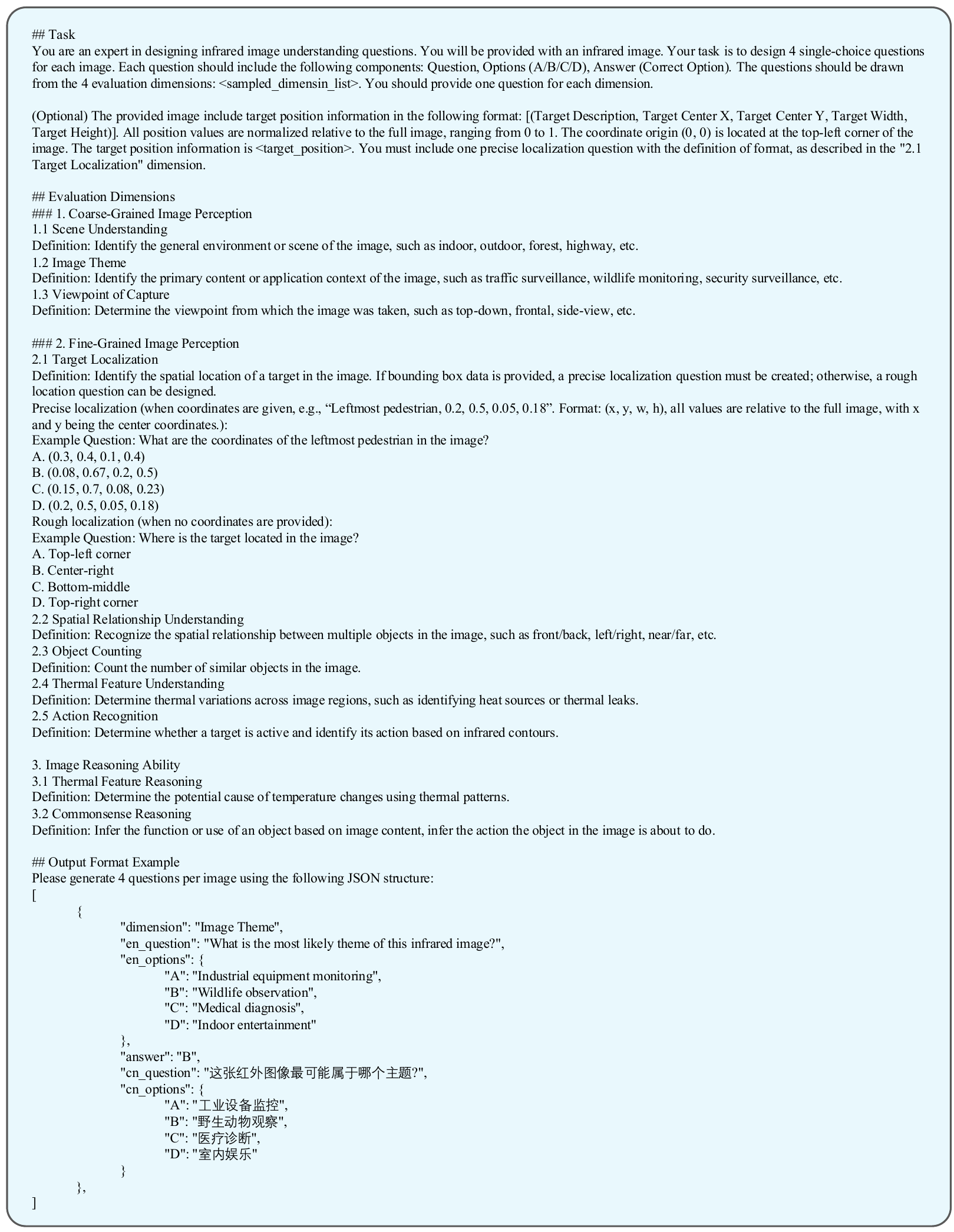}
\vspace{-3mm}
\caption{The system prompt used for VQA generation in \secref{sec:const_pipe}.}
\label{fig:qa_prompt}
\vspace{-3mm}
\end{figure*}

In this section, we provide the specific prompts used in this work. \figref{fig:eval_prompt} illustrates the unified system prompt when evaluating MLLMs on IF-Bench. \figref{fig:parse_prompt} depicts LLM-based parsing prompt, as described in \secref{sec:eval_protocol}. \tabref{fig:eval_prompt_dual} and \tabref{fig:textual_prior} provide the evaluation prompt for the dual-image input in GenViP and the textual prior, respectively. For simplicity, only the English versions are presented here, with the Chinese versions being semantically equivalent to their English counterparts. \figref{fig:qa_prompt} lists the system prompt used for automatic VQA generation in \secref{sec:const_pipe}.

\section{Details of Editing Models}
\label{sec:appendix_c}
In this section, we clarify the training and inference details involved in the editing models of GenViP. 

\textbf{Inference of Editing Models.} For Seedream 4.0 \cite{seedream4} and Gemini-2.5-Flash \cite{gemini2.5}, we use their official APIs for infrared-to-RGB translation. For Qwen-Edit-2509 \cite{qwen_edit}, inference is performed via local deployment. The number of inference steps is set to 40 for both Qwen-Edit-2509 and Qwen-Edit-2509-FT.

\textbf{Training of Qwen-Edit-2509.} We fine-tune Qwen-Edit-2509 \cite{qwen_edit} using LoRA \cite{lora} with a rank of 32. Qwen-Edit-2509-FT is trained for two epochs on the RGB-T dataset consisting of 50,000 pairs, with a batch size of 8 and a fixed resolution of 1024.

\textbf{Editing Prompt.} For both the training of Qwen-Edit-2509-FT and inference of all editing models, we use the same editing prompt: \textit{Translate the infrared image into the corresponding visible light (RGB) image}. We also experimented with more sophisticated editing prompts, but found that they led to similar results. Therefore, we adopt this simple prompt throughout.

\section{More Analyses}
\label{sec:appendix_d}
\begin{figure*}[p]
\centering
\includegraphics[width=0.98\linewidth]{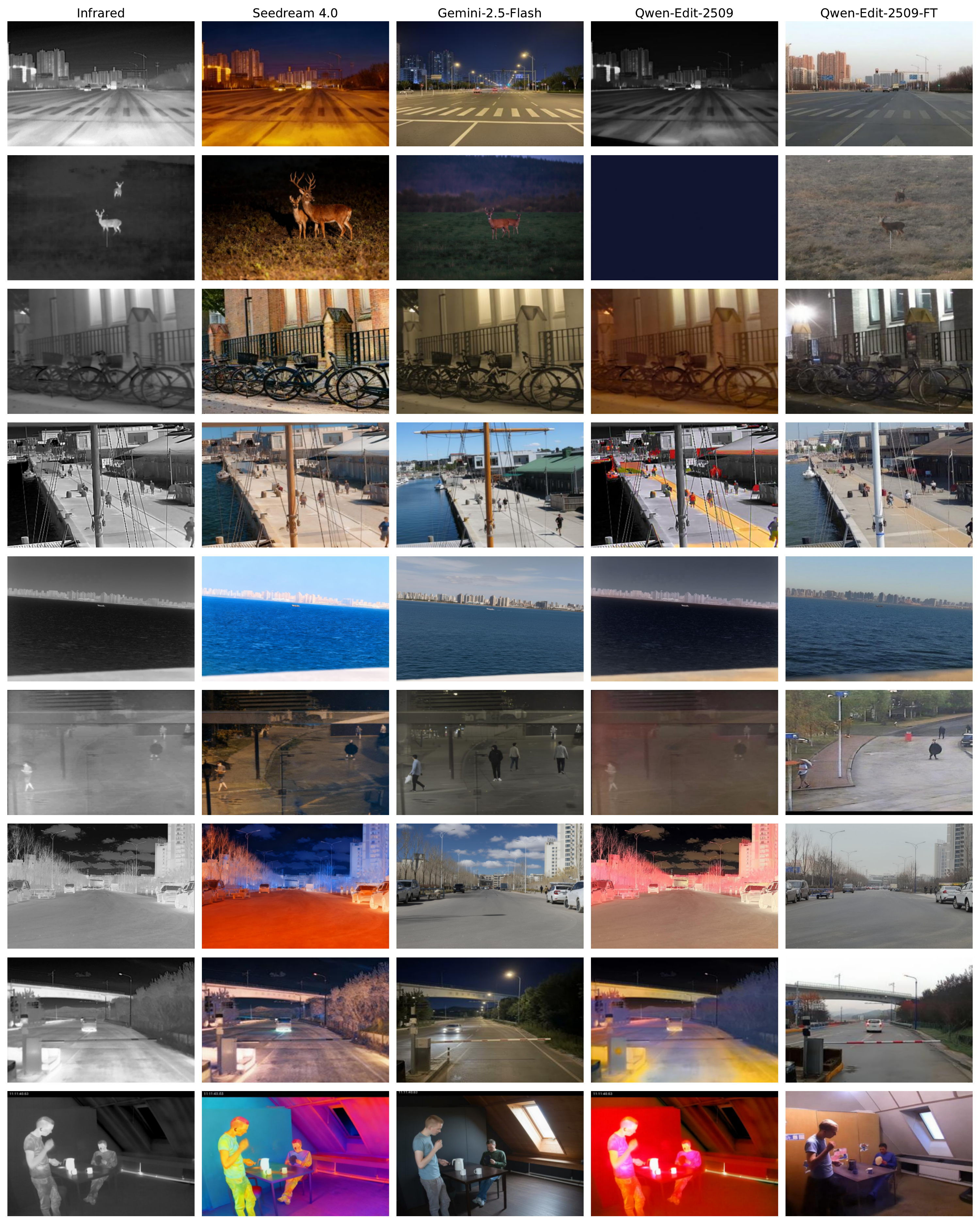}
\vspace{-3mm}
\caption{The visualization of translated RGB images from infrared ones across different editing models.}
\label{fig:translated_img}
\vspace{-3mm}
\end{figure*}
\textbf{Language Preference.} \figref{fig:language_pref} presents the Chinese and English scores of various MLLMs on IF-Bench. Most models exhibit no notable language preference, with score differences between the Chinese and English settings remaining within one point. Only a few smaller models, such as InternVL3.5-1B \cite{internvl3.5}, InternVL3.5-2B \cite{internvl3.5}, and Qwen3-VL-4B-Instruct \cite{qwen3_vl}, show slightly larger discrepancies of approximately one to two points.

\textbf{Stricter Correctness Judgement.} In the main paper, we adopt a circular evaluation strategy and report the average accuracy across different option orders. Here, we further apply a more stringent correctness criterion: a question is considered correct only if the model answers it correctly under all option orders. \tabref{tab:if_bench_strict} compares the score on IF-Bench of the two criteria. Notably, the stricter criterion substantially reduces accuracy, with the average score dropping by more than 10 points for several models. This indicates that current MLLMs exhibit limited robustness to option-order variations and still have considerable room for improvement in infrared image understanding.

\begin{table*}[ht]
\centering
\scriptsize
\setlength{\tabcolsep}{1.5mm}
\caption{Comparison of different correctness judgement strategies.}
 \vspace{-2mm}
\label{tab:if_bench_strict}
\begin{tabular}{lc|l|cccccccccc}
\toprule
Model & Stricter Strategy & \makecell[c]{Avg} & SU & IT & VC & TL & SRU & OC & TFU & AR & TFR & CR \\
\midrule
\multirow{2}{*}{Qwen2.5-VL-7B \cite{qwen2.5_vl}} & \xmark & \textbf{71.1} & 93.3 & 86.5 & 74.4 & 60.3 & 49.5 & 42.7 & 58.9 & 74.0 & 88.6 & 83.1 \\
& \cmark & 53.3 \textbf{\textcolor{ForestGreen}{(-17.8)}} & 89.7 & 71.6 & 55.2 & 10.3 & 31.5 & 17.8 & 43.6 & 58.5 & 77.1 & 77.4 \\
\midrule
\multirow{2}{*}{Qwen3-VL-8B-Instruct \cite{qwen3_vl}} & \xmark & \textbf{78.8} & 93.9 & 93.1 & 79.7 & 74.1 & 65.5 & 60.2 & 68.6 & 75.8 & 88.3 & 88.9 \\
& \cmark & 69.1 \textbf{\textcolor{ForestGreen}{(-9.7)}} & 91.7 & 89.2 & 69.0 & 54.8 & 51.9 & 40.8 & 61.4 & 66.2 & 81.4 & 84.7 \\
\midrule
\multirow{2}{*}{InternVL3.5-30B-A3B \cite{internvl3.5}} & \xmark & \textbf{74.4} & 94.9 & 88.2 & 73.7 & 75.0 & 59.0 & 45.0 & 69.0 & 68.1 & 85.8 & 85.7 \\
& \cmark & 58.8 \textbf{\textcolor{ForestGreen}{(-15.6)}} & 88.5 & 79.7 & 59.5 & 50.0 & 32.4 & 26.4 & 52.9 & 58.5 & 65.3 & 75.0 \\
\midrule
\multirow{2}{*}{InternVL3.5-38B \cite{internvl3.5}} & \xmark & \textbf{79.0} & 93.4 & 93.2 & 77.8 & 80.5 & 57.6 & 59.6 & 70.0 & 77.7 & 91.5 & 88.7 \\
& \cmark & 65.2 \textbf{\textcolor{ForestGreen}{(-13.8)}} & 89.1 & 80.4 & 66.4 & 62.3 & 43.5 & 29.3 & 55.7 & 62.3 & 80.5 & 82.3 \\
\midrule
\multirow{2}{*}{Qwen3-VL-235B-A22B-Instruct \cite{qwen3_vl}} & \xmark & \textbf{83.7} & 95.8 & 94.9 & 84.5 & 83.2 & 65.3 & 68.4 & 71.6 & 84.0 & 97.5 & 92.1 \\
& \cmark & 78.5 \textbf{\textcolor{ForestGreen}{(-5.2)}} & 94.9 & 90.5 & 75.9 & 74.0 & 57.4 & 59.2 & 67.1 & 80.0 & 96.6 & 89.5\\

\bottomrule
\end{tabular}
\end{table*}

\textbf{Relationship with Thinking-with-Image.} Thinking-with-Image was introduced in OpenAI o3 \cite{gpto3}, where the model performs a variety of operations on the input image to actively generate intermediate visual representations, thereby enhancing its capability in multimodal reasoning. Our proposed GenViP shares a similar intuition: processing the image to improve the model’s understanding. But the key difference is that in our work, the model is explicitly required to invoke an editing model at the initial stage. The idea behind GenViP can be naturally extended to the Thinking-with-Image paradigm, where the model could be trained to autonomously decide whether to call an editing model. We leave this direction for future work.

\textbf{Visualization of Translated Quality.} \figref{fig:translated_img} illustrates the translated RGB images produced by different editing models. We observe that for some infrared inputs, closed-source models such as Seedream 4.0 \cite{seedream4} and Gemini-2.5-Flash \cite{gemini2.5} show strong generalization ability (e.g., rows 1, 3, and 5). However, their translation quality remains suboptimal for certain other cases, exhibiting issues such as unnatural color rendering and imperfect spatial correspondence. Qwen-Edit-2509 \cite{qwen_edit} performs worse overall compared to the closed-source models. After fine-tuning on 50,000 infrared–RGB image pairs, Qwen-Edit-2509-FT significantly improves the naturalness of the translated images as well as their spatial and semantic consistency, thereby leading to a substantial performance gain for GenViP.
